\documentclass[12pt,onecolumn]{IEEEtran}
\usepackage{setspace}
\usepackage{comment}
\usepackage{changepage}
\usepackage{amsmath}
\usepackage{amssymb}
\usepackage[dvips]{graphicx}
\usepackage{setspace}
\usepackage{epsfig}
\usepackage{color}
\usepackage{morefloats}
\usepackage{enumerate}
\usepackage{bbm}
\usepackage{cases}
\usepackage{enumitem}
\usepackage{cite}
\usepackage{graphicx}

\allowdisplaybreaks
\newcommand{\mtA}{{\mathcal{A}}}
\newcommand{\mtS}{{\mathcal{S}}}

\newcommand{\PL}{{\rm{PL}}}
\newcommand{\EE}{{\rm{EE}}}

\newcommand{\cmark}{\checkmark}

\usepackage{algorithm}
\usepackage{algpseudocode}
\usepackage{amsmath}

\usepackage{comment}
\usepackage{caption}
\usepackage{subcaption}

\newcommand{\gNB}{\rm{gNB}}
\usepackage{dsfont}
\usepackage[a4paper,
            bindingoffset=0.2in,
            left=0.6in,
            right=0.6in,
            top=0.7in,
            bottom=0.9in,
            footskip=.25in]{geometry}
\usepackage{booktabs}   

\usepackage{tabularx}
\usepackage{multirow}
\usepackage{array}
\usepackage{makecell}
\usepackage{graphicx} 

\newcolumntype{P}[1]{>{\RaggedRight\arraybackslash}p{#1}}
\newcolumntype{C}[1]{>{\centering\arraybackslash}p{#1}}

\newcolumntype{Y}{>{\raggedright\arraybackslash}X} 

\begin{document}

\title{Energy-Efficient Sleep Mode Optimization in 5G mmWave Networks via Multi-Agent Deep Reinforcement Learning}
\author{Ismail Guvenc\vspace{-0.5cm}}
\author{Saad Masrur$^1$, \.{I}smail G\"{u}ven\c{c}$^1$, David L\'opez-P\'erez$^2$\\
$^1$Department of Electrical and Computer Engineering, North Carolina State University, Raleigh, NC\\
$^2$Universitat Polit\`ecnica de Val\`encia, Valencia, Spain\\
{\tt \{smasrur,iguvenc\}@ncsu.edu}
\thanks{This research is supported in part by the NSF under award numbers CNS-2332835 and CNS-1910153, the Generalitat Valenciana through the CIDEGENT PlaGenT, Grant CIDEXG/2022/17, Project iTENTE, and by the action CNS2023-144333, financed by MCIN/AEI/10.13039/501100011033 and the European Union “NextGenerationEU”/PRTR.”}
}




\vspace{-0.4cm}
\maketitle
\thispagestyle{empty}

\begin{abstract}
\pagestyle{empty}



Dynamic sleep mode optimization (SMO) in millimeter-wave (mmWave) networks is essential for maximizing energy efficiency (EE) under stringent quality-of-service (QoS) constraints. However, existing optimization and reinforcement learning (RL)-based approaches rely on aggregated, static base station (BS) traffic models that fail to capture non-stationary traffic dynamics and suffer from prohibitively large state–action spaces, limiting their real-world deployment.
To address these challenges, this paper proposes a Multi-Agent Deep Reinforcement Learning (MARL) framework employing a Double Deep Q-Network (DDQN), referred to as \texttt{MARL-DDQN}, for adaptive SMO in a 3D urban environment using a time-varying and community-based user equipment (UE) mobility model. Unlike conventional single-agent RL, the proposed \texttt{MARL-DDQN} enables scalable, distributed decision-making with minimal signaling overhead. A realistic BS power consumption model and beamforming are integrated to accurately quantify EE, while QoS is uniquely defined in terms of throughput.  The proposed method adaptively learns SMO policies to maximize EE while mitigating inter-cell interference and ensuring throughput fairness. Extensive simulations demonstrate that \texttt{MARL-DDQN} consistently outperforms state-of-the-art SM strategies, including the \texttt{All On}, iterative QoS-aware load-based (\texttt{IT-QoS-LB}), \texttt{MARL-DDPG}, and \texttt{MARL-PPO}, achieving up to $0.60$ Mbit/Joule EE, $8.5$ Mbps  10th-percentile throughput, and satisfying QoS constraints $95$\% of the time under dynamic network scenarios.






\pagestyle{empty}

\textit{Index~Terms}--- Energy efficiency, mmWave, multi-agent reinforcement learning, sleep mode optimization.
\end{abstract}



\vspace{-0.4cm}

\section{Introduction}\label{sec:intro}
\pagestyle{empty}

The exponential growth in mobile data demand has necessitated increased spectrum availability and accelerated the expansion of cellular network infrastructure. To address the limitations of the sub-6 GHz spectrum, millimeter wave (mmWave) communications, operating within the 30–300 GHz band, have emerged as a key enabler in fifth-generation (5G) networks. With significantly larger bandwidth availability, mmWave technology presents a viable solution to spectrum scarcity challenges \cite{baldemair2015ultra}. However, mmWave signals suffer from high propagation loss, atmospheric absorption, and susceptibility to blockages, which severely limit coverage and reliability. 

To address coverage and growing capacity demands, 5G networks rely on densification, deploying numerous low-power mmWave BSs with inter-site distances of a few hundred meters \cite{baldemair2015ultra}. These BSs utilize large antenna arrays to enable beamforming and spatial multiplexing, often relying on hybrid analog-digital precoding to reduce hardware complexity \cite{alkhateeb2015limited}. However, the RF chain remains a major source of power consumption, particularly the Analog-to-digital converters (ADCs) and digital-to-analog converters (DACs), whose power scales with sampling rate. Due to the higher frequencies and wider bandwidths of mmWave systems, these components require significantly higher sampling rates than sub-6 GHz systems \cite{feng2018dynamic}, resulting in substantial energy demands. Consequently, dense deployments substantially increase energy consumption, posing sustainability and operational efficiency challenges. As a result, improving energy efficiency (EE) has become a key focus for researchers and operators to enable sustainable network deployment. As the demand for higher data rates continues to grow, 6G networks will further enhance spectral efficiency (SE) and system capacity through ultra-dense network architectures, making energy-efficient network operation even more critical.

In cellular networks, BSs account for 60\% to 80\% of total energy consumption, and interestingly, BS traffic load is less than a tenth of the peak value for 30\% of the time on weekdays \cite{oh2011toward}. Dynamic Sleep Mode Optimization (SMO) offers a promising way to lower this energy footprint by selectively activating or deactivating BSs based on instantaneous demand. However, careless deactivation can degrade coverage, reduce throughput, and violate quality-of-service (QoS) guarantees, making it essential to jointly balance energy efficiency (EE) and user experience. Achieving this balance is challenging because user mobility, spatiotemporal traffic variation, and fluctuating channel conditions evolve rapidly, requiring adaptive, data-driven policies that can learn optimal BS operation under dynamic network environments.


While numerous studies on SMO in BS networks have focused on improving EE, many of these studies overlook critical factors such as UE distribution modeling, QoS requirements in terms of throughput, mutual interference among BSs, and accurate power modeling, key components in EE analysis. Additionally, several studies have incorporated load-dependent BS power models. However, most have focused on sub-6 GHz networks with simplified or aggregated traffic representations. In contrast, mmWave systems with large antenna arrays and hybrid beamforming exhibit distinct RF-chain and hardware power dynamics, necessitating customized power modeling.

Reinforcement Learning (RL) has emerged as a promising framework for sequential decision-making in dynamic and complex environments. Existing Deep RL (DRL)-based approaches for SMO have demonstrated energy-saving potential, yet most are constrained to small-scale scenarios because the joint action space grows exponentially with the number of BSs. This limitation becomes especially critical in large-scale dense network scenarios, where the number of BSs makes conventional single-agent DRL approaches impractical. To overcome this challenge, we formulate SMO as a Markov Decision Process (MDP) and develop a Multi-Agent Deep Reinforcement Learning (MARL) framework based on Double Deep Q-Networks (DDQN) referred as \texttt{MARL-DDQN}. In this design, each BS operates as an independent agent that learns adaptive activation policies, enabling scalable and stable training across high-dimensional environments while maintaining a balance between EE and user throughput.



Beyond scalability, realistic modeling of mmWave-specific characteristics is vital for learning meaningful SMO policies. In mmWave networks, spatio-temporal user dynamics and beamforming behavior directly affect both network performance and power consumption. Rapid variations in UE distribution cause fluctuating BS loads, while directional beamforming alters hardware activity, specifically RF-chain usage and power amplifier (PA) efficiency, thereby impacting total power draw. To capture these effects, our framework explicitly integrates dynamic UE distribution, beamforming-aware modeling, and 3D propagation environments to emulate near-real-world conditions. The 3D urban model enables a realistic representation of the physical environment, capturing real-life structural layouts and accurately modeling LOS/NLOS transitions and blockage effects critical to dense mmWave deployments. To the best of our knowledge, this is the first study to integrate 3D geometry with MARL-driven sleep-mode optimization for mmWave networks. The proposed solution operates in two phases: (i) selecting an optimal subset of BSs within the 3D environment to maximize coverage, and (ii) developing an adaptive SMO strategy that maximizes EE while satisfying UE throughput requirements.

The key contributions of this paper include the following:
\begin{itemize}
    \item We propose a \texttt{MARL-DDQN} framework for SMO, addressing the scalability limitations of existing DRL-based methods. The proposed design decomposes the learning process across multiple BS agents, enabling scalable and stable training. The framework also exploits state information extracted via UE clustering to refine agent decisions, ensuring efficient exploration of large state–action spaces and robust adaptation to varying network dynamics.
    \item We incorporate a time-variant community-based UE mobility model in a realistic 3D urban macro (UMa) environment to accurately capture real-world dynamics in mmWave communications.
    \item We develop a realistic power consumption model for mmWave BSs to enable a more precise EE analysis. The proposed framework maximizes EE while incorporating inter-BS interference, ensuring QoS constraints in terms of throughput, and leveraging beamforming techniques within the SMO framework to improve the mmWave link budget.
    \item We evaluate the performance of the proposed \texttt{MARL-DDQN} algorithm by benchmarking it against MARL proximal policy optimization (\texttt{MARL-PPO}), MARL deep deterministic policy gradient (\texttt{MARL-DDPG}), iterative QoS-aware load-based (\texttt{IT-QoS-LB})~\cite{celebi2019load}, and the \texttt{All On} strategies.

    \item We conduct an extensive evaluation of the proposed \texttt{MARL-DDQN} algorithm, analyzing the impact of dynamic SMO on EE, throughput, and QoS while varying the number of BSs and UEs in the network to assess its scalability and performance under different deployment scenarios.
\end{itemize}

The subsequent sections of the paper are structured as follows. Section II provides a comprehensive review of the existing literature. Section III introduces the system model, while Section IV outlines the problem formulation, discusses state-of-the-art (SOTA) SMO strategies, and provides a brief overview of MARL for SMO. Section V describes the proposed \texttt{MARL-DDQN} algorithm in detail. Section VI presents the numerical results and performance evaluation. Finally, Section VII concludes the paper.
\section{Literature Review}
Despite extensive research on SMO and BS on/off scheduling, a comprehensive study addressing SMO in mmWave networks under diverse BS and UE distributions remains limited. Optimization-based and DL/RL-based approaches have been explored for dynamic SMO, where sleep-cycle durations range from seconds to hours depending on network dynamics \cite{oh2011toward, celebi2019load}. However, many existing works such as \cite{saker2010minimizing, malta2023using, liu2015small, guo2013optimal} assume a well-defined network with mathematically tractable models, often relying on Poisson traffic assumptions, while these models provide insights into system behavior, they fail to capture the complex real-world dynamics. They often assume precise knowledge of network conditions and UE locations, assumptions that rarely hold due to the inherently stochastic nature of real-world wireless environments. The aforementioned studies, along with \cite{saker2010minimizing, peng2014greenbsn, onireti2017analytical}, primarily consider aggregated traffic at the BS and fail to account for mutual interference among BSs, which is a crucial factor in network performance. Some studies, such as \cite{masrur2024energy, li2015optimization, ju2022energy}, do not consider UE QoS requirements, while others rely on latency as a QoS metric. However, latency is highly variable, influenced by factors such as network congestion, interference, and backhaul delays, making it an unreliable sole indicator of QoS.

Several studies have proposed SMO techniques tailored for different network settings. For instance, \cite{garcia2020energy} introduced energy-efficient SM techniques for cell-free mmWave massive MIMO networks using a log-normal traffic map for UE distribution, improving tractability but failing to reflect realistic mobility and traffic dynamics. Similarly, \cite{kim2023optimization} used binary integer programming for SMO but assumed static UE locations and a load-independent BS power model, overlooking mobility and realistic power dynamics. Queueing theory has been widely used to balance EE and QoS degradation \cite{pervaizenergy, onireti2017analytical, guo2013optimal}. For example, \cite{pervaizenergy} adopted a simplified BS power model but, like many others, did not account for UE mobility. Instead, it relied on aggregated BS traffic, abstracting away UE-level details and overlooking interference variations caused by mobility. Similarly, \cite{rengarajan2015energy, saker2010minimizing}, modeled aggregate traffic using a Poisson process, which fails to reflect real-world spatial and temporal traffic fluctuations, where UEs generate highly variable and bursty traffic. In \cite{liu2015small}, a threshold-based SMO algorithm deactivates small cells when traffic falls below a set threshold, assuming a Poisson traffic model.

DL and RL have been widely applied to optimize wireless communication systems, including SMO \cite{maghsudi2016multi}, \cite{masrur2024energy}. In our previous work \cite{masrur2024energy}, we proposed a contextual deep Multi-Armed Bandit (MAB)-based SMO algorithm, assuming randomly distributed UEs. However, this approach did not account for UE QoS and relied on a fixed BS power model, independent of network load. Additionally, due to the inherent limitations of MAB, scalability remained a significant challenge. In \cite{ye2019drag} and \cite{wu2021deep}, deep reinforcement learning (DRL)-based approaches put a subset of BSs into SM to reduce energy consumption. However, these works assume aggregated traffic at the BS and do not model UE mobility. They rely on traffic forecasting models, making SMO decisions highly sensitive to errors in arrival rate predictions. Moreover, BS placement is not optimized. Additionally, \cite{ye2019drag} lacks EE performance evaluation. Traffic prediction-based SMO has also been explored. Long Short-Term Memory (LSTM)-based traffic prediction was used in \cite{zhu2021joint}, where BS SMO was formulated as a mixed-integer programming problem. Similarly, \cite{hoffmann2021increasing} applied Q-learning using a radio environment map with UE locations, bit rates, and power data. However, the centralized design incurs high signaling overhead, and the BS power model ignores load variations, limiting practicality and accuracy.

Q-learning was utilized in \cite{el2019location} to select different SMs, where BSs without active UEs were put into SM. However, this scenario is unrealistic, as completely idle BSs are rare in real-world networks. Moreover, UE mobility was not considered. Similarly, \cite{ju2022energy} employed DRL for SMO with UEs moving at random velocities but did not incorporate QoS constraints. Additionally, the study only considered four UEs and ten small cells, which does not accurately represent large-scale real-world deployments.  Other RL-based works, such as \cite{masoudi2020reinforcement,malta2023using}, rely on policy-based RL algorithms but assume fixed UE locations and aggregated traffic patterns, ignoring dynamic mobility and real-time traffic variability.  Beyond single-agent RL, MARL has also been explored. In \cite{el2020reinforcement}, BSs make SMO decisions based on queue-based UE requests, assuming static UE locations. Similarly, \cite{el2019distributed} adopts an aggregated traffic model, disregarding individual UE QoS.

\begin{table*}[t]
\centering
\caption{Representative BS SMO studies and the gaps addressed by this work.}
\tiny
\setlength{\tabcolsep}{2pt}
\begin{tabular}{@{} l c l c c c c c @{}}
\toprule
\textbf{Study} & \textbf{Type} &
\makecell{\textbf{Traffic Model}\\(Agg.\ vs UE-level)} &
\makecell{\textbf{QoS Metric}\\(Thrpt/Lat/None)} &
\makecell{\textbf{Power Model}\\(Static / Load-dep)} &
\textbf{Interf.} & \textbf{3D/mmWave} &
\makecell{\textbf{Scalability}\\(Low$\rightarrow$High)}\\
\midrule
\cite{saker2010minimizing} & Opt./Heur. & Aggregated & × & Static & × & × & Low\\
\cite{guo2013optimal}      & Opt./Heur. & Aggregated & Lat & Static & × & × & Low\\
\cite{liu2015small}        & Heur.      & Poisson/Agg. & × & Static & × & × & Low\\
\cite{onireti2017analytical} & Analytical & Aggregated & × & Static & × & × & Low\\
\cite{pervaizenergy}       & Analytical & Aggregated & Lat & Static & × & × & Low\\
\cite{rengarajan2015energy} & Heur.      & Poisson/Agg. & × & Load-dep & × & × & Low\\
\cite{garcia2020energy}    & Heur.      & Log-normal map & × & Static & \cmark & \cmark & Lo\\
\cite{kim2023optimization} & BIP        & Static UE & × & Static & \cmark & × & Low\\
\midrule
\cite{zhu2021joint}        & MIP–DL     & Aggregated & × & Static & \cmark & × & Low\\
\cite{masrur2024energy}    & MAB        & Random UE-level & × & Static & × & × & Low\\
\cite{hoffmann2021increasing} & QL (SA)  & Aggregated & Thrpt & Static & \cmark & × & Low\\
\cite{el2019location}      & QL (SA)    & UE-level & × & Static & × & × & Low\\
\cite{ye2019drag}          & DRL (SA)   & Aggregated & Lat & Load-dep & \cmark & × & Medium\\
\cite{wu2021deep}          & DRL (SA)   & Forecast/Agg. & Lat & Load-dep & \cmark & × & Medium\\
\cite{ju2022energy}        & DRL (SA)   & Uniform & Thrpt & Load-dep & \cmark & × & Low\\
\cite{el2020reinforcement} & MARL       & Aggregated & Lat & Static & × & × & Low\\
\cite{el2019distributed}   & MARL       & Aggregated & Lat & Static & × & × & Low\\
\midrule
\textbf{Ours} & \textbf{MARL} & \textbf{UE-level + mobility} &
\textbf{Thrpt} & \textbf{Load-dep.} & \textbf{\cmark} &
\textbf{\cmark (3D UMa)} & \textbf{Med.\,$\rightarrow$\,High}\\
\bottomrule
\end{tabular}
\vspace{1mm}

\raggedright\tiny
\textbf{Notes:} “×” \(=\) not considered; “\cmark” \(=\) considered.  
“Opt.” \(=\) optimization-based, “Heur.” \(=\) heuristic;  
“Thrpt” \(=\) throughput, “Lat” \(=\) latency, “Agg.” \(=\) aggregated;  
“Load-dep.” \(=\) load-dependent power model.
\label{tab:related_work}
\end{table*}

As summarized in Table~\ref{tab:related_work}, most prior SMO studies adopt aggregated traffic models without explicit UE mobility, rely on static or simplified BS power models, and rarely incorporate mmWave-specific considerations such as beamforming-aware power or 3D blockage. In contrast, our MARL framework integrates UE-level dynamics, throughput-based QoS, interference coupling, and an mmWave load-dependent, beamforming-aware power model within a 3D urban environment, while remaining scalable to dense deployments.


\vspace{-0.1cm}

\section{System Model}\label{sec:system}
\begin{figure}
\vspace{-0.5cm}
\includegraphics[width=0.6\linewidth]{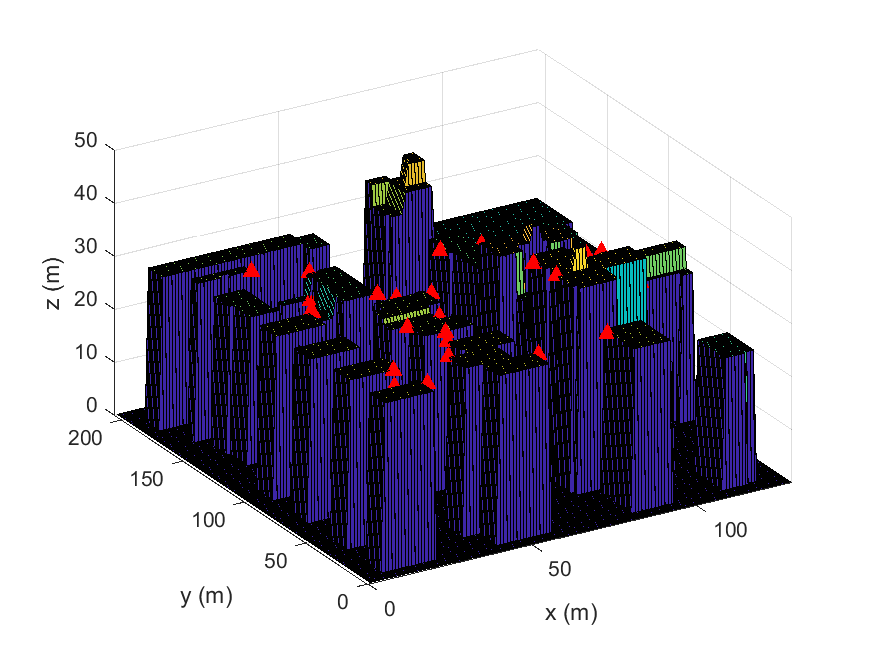}
	\centering
\vspace{-0.3cm}
	\caption{Urban Macro (UMa) outdoor-to-outdoor communication scenario, where the reduced BS positions \eqref{RBS} are represented by red triangles.}
	\label{fig1}
\vspace{-0.35cm}
\end{figure}
Our study is based on  UMa outdoor-to-outdoor communication scenario. This scenario incorporates various elements such as terrain data, landmarks, and street routes, all of which are illustrated in Fig. \ref{fig1}. Note that  3D geographic data of some real environments can be obtained from publicly available sites such as Open-Street-Map \cite{haklay2008openstreetmap}, to characterize real-world scenarios. Using this data, we have constructed a 3D map with a digital elevation model. This map is structured as a grid, where each square contains data about the location and height of its central point. The model includes buildings of varying heights and widths selected within the ranges of 8-25 meters and 20-45 meters, respectively. The BSs are positioned on the rooftops, while the UEs are located on the ground. We focused on outdoor communication due to the nature of cellular mmWave networks, therefore, we have defined the service area (SA) as the total area excluding the area covered by the buildings. This approach ensures that our focus remains on the areas that require service.
\subsection{BS Placement}
The \(N_c\) candidate BS locations, denoted by \(\mathcal{P}_{\rm c}^{\mathrm{BS}}\), correspond to building boundaries facing the SA. Given the large \(N_c\), deploying BSs at all sites is impractical, and an excessive number of BSs would not reflect realistic, strategically planned deployments. In our study, we first strategically minimized the BS candidate locations in a UMa scenario. We used an iterative algorithm that selects the BS with the maximum visibility. This choice follows mmWave deployment studies showing that, due to the quasi-optical nature of propagation at 28–73 GHz, link feasibility mainly depends on line-of-sight and reflection visibility, while interference and noise have secondary effects once a path exists \cite{anjinappa2021tvt}. Visibility refers to the extent to which a grid is observable from a given location, which is calculated using the voxel viewshed algorithm \cite{messerli2015image}. The iterative algorithm adds the BS that improves visibility the most in each iteration. This process continues until no further visibility improvement is possible, resulting in a reduced set of BSs that optimally covers the whole SA, and the reduced collection of BSs positions is denoted by the set:
\begin{equation} \label{RBS}
\mathcal{P}_{\rm r}^{\mathrm{BS}} \in\left\{\left(x_i^{\mathrm{BS}}, y_i^{\mathrm{BS}}, z_i^{\mathrm{BS}}\right) \mid \forall i \in[N_{\rm r}]\right\}~,
\end{equation}
where $N_{\rm r}$ (i.e., $N_{\rm r} \ll N_{\rm c}$) represents the total number of BS locations that are capable of covering the entire SA. The SA is split into $M$ grid points where the UEs can be located, and the positions of these points are defined as follows:
\begin{equation} \label{PCUE}
\mathcal{P}^{\mathrm{SA}} \in\left\{\left(x_i^{\mathrm{SA}}, y_i^{\mathrm{SA}}, z_i^{\mathrm{SA}}\right) \mid \forall i \in[M]\right\}~.
\end{equation}

\begin{figure}
\vspace{-0.2cm}
	\includegraphics[width=0.6\linewidth]{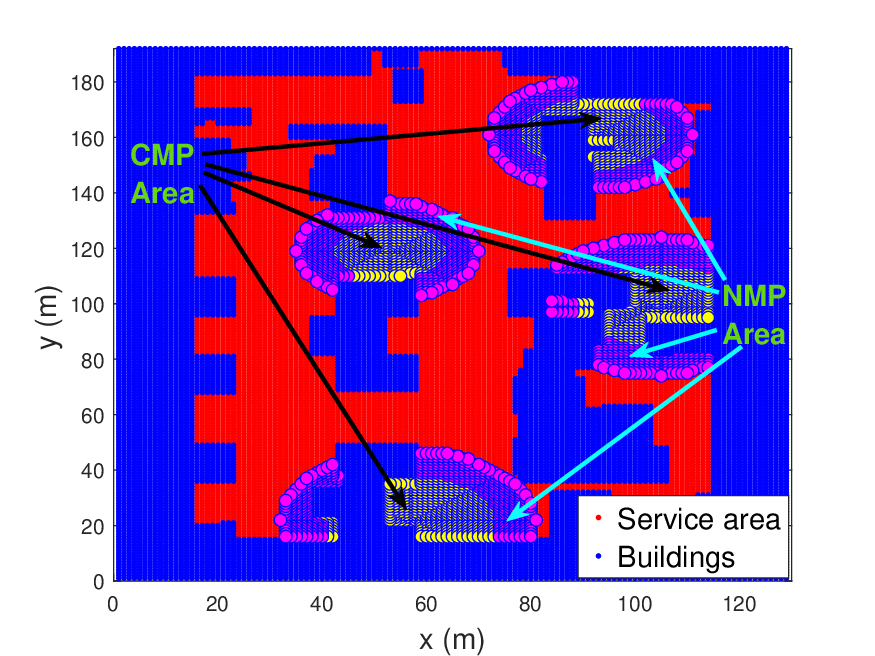}
	\centering
\vspace{-0.3cm}
	\caption{ Illustration of the NMP and CMP areas. The SA is outlined in red, while buildings are represented in blue. A total of \(C = 4\) communities are depicted. The CMP area is highlighted in yellow, whereas the NMP area encompasses both the yellow and magenta regions.}
	\label{NMPCMP}
\vspace{-0.35cm}
\end{figure}
\vspace{-0.25cm}
\subsection{UE Mobility Model} \label{mobility}

Most existing works model UE traffic using aggregated or independently and identically distributed (i.i.d.) assumptions, which simplify analysis but overlook spatial–temporal mobility observed in real networks. To address this, we adopt the time-varying community-based mobility model \cite{hsu2007modeling}, which captures skewed location preferences and the periodic reappearance of UEs at frequently visited sites, reflecting realistic mobility behavior. The time-varying community model consists of two distinct periods: Normal Movement Period (NMP) and Concentrated Movement Period (CMP), with fixed durations \(T_{\rm n}\) and \(T_{\rm c}\), respectively. The total duration of an episode is \(T_{\rm e} = T_{\rm n} + T_{\rm c}\), where each episode recurs periodically. 

During each period (\(T_{\rm n} \text{ or } T_{\rm c}\)), UEs are assigned to predefined communities (\(C\)), which represent frequently visited locations. Community sizes vary depending on the movement period: during NMP, communities cover an area of \(A_{\rm n}\)~\(\rm m^2\), while in CMP, the community size is reduced to \(A_{\rm c}\)~\(\rm m^2\) (\(A_{\rm c} < A_{\rm n}\)). Communities are randomly distributed within the SA, where \(C\) random grid points are selected, around which communities are defined in a circular fashion with areas \(A_{\rm n}\) and \(A_{\rm c}\). The classification of community areas for NMP and CMP is illustrated in Figure 2. 



UEs are randomly assigned to \( C \) communities, evenly when divisible; otherwise, some communities hold fewer UEs. Within each time period, either NMP or CMP, UE movement follows two distinct modes: local epoch and roaming epoch. During the local epoch, the UE movement is confined within its assigned community, while in the roaming epoch, the UE can traverse the entire SA. At the beginning of each epoch, the UE selects a speed uniformly from \([v_{\rm min}, v_{\rm max}]\) and an angle uniformly from \([0, 2\pi]\). If a UE reaches the boundary of the SA during an epoch, it is reflected within the valid region based on the mobility constraints of the current epoch \cite{7784849}.

\begin{figure}
\vspace{-0.5cm}
	\includegraphics[width=0.5\linewidth]{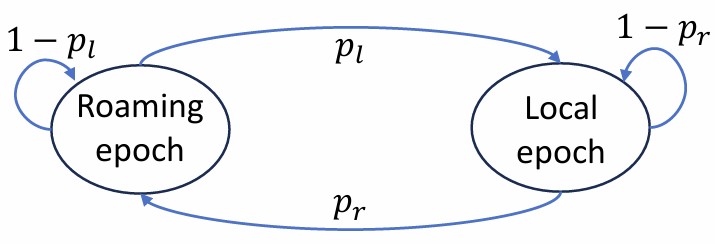}
	\centering
\vspace{-0.3cm}
	\caption{Two-state Markov model governing UE transitions between the roaming and local epochs.}
	\label{Twostates}
\vspace{-0.35cm}
\end{figure}

Each period consists of \(N_E\) epochs, with the duration of each epoch drawn from an exponential distribution with a mean of \(\bar{E}\). To ensure consistency, epoch durations are adjusted such that their cumulative length matches the total duration of the corresponding period. At the end of each epoch, UEs transition between movement modes according to a two-state Markov model, as illustrated in Fig.~ \ref{Twostates}. If a UE is in the roaming epoch, it remains in this mode with probability \(1 - p_{\rm l}\) or transitions to a local epoch with probability \(p_{\rm l}\). The transition probabilities differ between NMP and CMP, denoted as \(p_{\rm l,n}\) and \(p_{\rm l,c}\), respectively, where \(\rm n\) and \(\rm c\) correspond to NMP and CMP.

\vspace{-0.25cm}
\subsection{Bandwidth Allocation and Throughput Calculation}

The reference signal received power (RSRP) of UE $i$ from BS $j$ can be expressed as:
\begin{equation}
P^{\mathrm{rcv}}_{i,j}=P^{\mathrm{tx}}_{j}+ \Upsilon_{i,j}-\PL_{i,j}~,
\end{equation}
where $P^{\mathrm{tx}}_{j}$ denotes the transmit power of $j^{th}$ BS, $\Upsilon_{i,j}$ represents the directivity gain due to beamforming, and $\PL_{i,j}$ is the path loss between UE $i$ and BS $j$. Each BS employs codebook-based analog beamforming using a precomputed steering matrix (\texttt{SteeringMat}) that defines the array’s directional radiation pattern over 13 discrete azimuth and 13 elevation beam directions (169 beams in total). For each UE, the BS selects the beam that maximizes the array response toward the UE’s azimuth–elevation direction \((\theta_i,\phi_i)\). The corresponding beamforming gain is
\begin{equation}
\Upsilon_{i,j}=10\log_{10}\big(|\mathbf{a}^H(\theta_i,\phi_i)\mathbf{w}_b|^2\big)~,
\end{equation}
where \( \mathbf{a}(\theta_i,\phi_i) \) and \( \mathbf{w}_b \) are the array steering and beamforming vectors, respectively. The steering matrix thus serves as a lookup table for evaluating \( \Upsilon_{i,j} \) across all possible BS–UE angular combinations.

The path loss models for the UMa scenario, considering both line of sight (LOS) and non-line of sight (NLOS) conditions, have been utilized as outlined in the 3GPP TR 38.901 technical report \cite{zhu20213gpp}:
\begin{equation}
\PL_{i, j} = \left\{\begin{array}{l}
28.0 + 20\log_{10}(f_{\rm o}) + 22\log_{10}(d^{3D}_{i,j}), \text { for LOS } \\
32.4 + 20\log_{10}(f_{\rm o}) + 30\log_{10}(d^{3D}_{i,j}), \text { for NLOS }
\end{array}\right.\nonumber 
\end{equation}
where $d^{3D}_{i,j}$ is the 3D distance (m) between the BS $j$ and the UE $i$, and $f_{\rm o}$ is the center frequency.

We defined binary variables $u_{i,j}$, and $s_{i,j}$ to represent the UE association to BS, and the presence of a UE within the SA of other BSs except the serving BS, respectively:

\noindent
\begin{minipage}{0.43\textwidth}
\begin{equation}
\begin{aligned}
u_{i, j} &\doteq 
\begin{cases}
1, & \text{UE } i \text{ served by } \mathrm{BS}\ j \\
0, & \text{otherwise}
\end{cases}
\end{aligned}
\label{association}
\end{equation}
\end{minipage}
\hspace{0.2mm}
\begin{minipage}{0.52\textwidth}
\begin{equation}
\begin{aligned}
s_{i, j} &\doteq 
\begin{cases}
1, & \text{UE } i \text{ in SA of } \mathrm{BS}\ j, u_{i, j} \neq 1 \\
0, & \text{otherwise}
\end{cases}
\end{aligned}
\end{equation}
\end{minipage}

where $i =1,2, \ldots, U, j=0,1, \text{and} \ldots, N$ (i.e., $U\leq M$, $N\leq N_{\rm r}$). Here, $U$ is the total number of UEs that can be located anywhere within the SA, as defined by the set $\mathcal{P}^{\mathrm{SA}}$ \eqref{PCUE}. On the other hand, $N$ denotes the number of BSs whose locations are selected from the reduced set $\mathcal{P}_{\rm r}^{\mathrm{BS}}$ \eqref{RBS} ($\mathcal{P}_{\rm r}^{\mathrm{BS}} \subseteq \mathcal{P}_{\rm c}^{\mathrm{BS}}$). The binary variable $s_{i, j}$ quantifies interference from non-serving, nearby BSs.  UEs are associated with the BS providing the highest RSRP, rather than based on the closest distance to the BS. This approach ensures a more realistic network association, as RSRP accounts for path loss, antenna gains, and BS transmission power, unlike distance-based association, which may lead to suboptimal links.

Each BS is allocated a bandwidth (BW) of 50 MHz. Assuming a subcarrier spacing of 120 kHz, the system comprises 416 subcarriers, resulting in a total of \(N_{\text{PRB}} = 34\) Physical Resource Blocks (PRBs) per BS. The total available PRBs in the network are: \(
N_{\text{T-PRB}} = N_{\text{PRB}} \times N
\). To ensure fair resource allocation, we define an upper limit on the number of PRBs assigned to each UE as follows:
\begin{equation}
    N^{\rm max}_{\text{PRB}, \rm UE} = \lfloor \frac{N_{\text{T-PRB}}}{U} \rfloor~.
\end{equation}
Since the UE distribution across BSs is non-uniform, some BSs may have a higher UE density, making it infeasible to allocate \( N^{\rm max}_{\text{PRB}, \rm UE} \) to every UE. To address this, we employ an adaptive PRB allocation scheme, where every UE associated with the $j$-th BS will have \(U_{\text{PRB},j}\) PRBs:  
\begin{equation}
    U_{\text{PRB},j} =
\begin{cases} 
N^{\rm max}_{\text{PRB}, \rm UE} , & \text{if }  N^{\rm max}_{\text{PRB}, \rm UE} \times N_{\text{UE},j} \leq N_{\rm PRB}  \\
\lfloor \frac{N_{\text{PRB}}}{N_{\text{UE},j}} \rfloor, & \text{if } N^{\rm max}_{\text{PRB}, \rm UE} \times N_{\text{UE},j} > N_{\rm PRB} 
\end{cases}~, \label{eq:PRB}
\end{equation}
where \( N_{\text{UE},j} \) denotes the number of UEs associated with BS \( j \). If a BS is overloaded (i.e., case 2 in \eqref{eq:PRB}), a round-robin scheduling mechanism is employed to allocate PRBs proportionally among UEs. Otherwise (i.e., case 1 in \eqref{eq:PRB}) a UE will have \(N^{\rm max}_{\text{PRB}, \rm UE}\) PRBs. This approach ensures equitable resource distribution while maintaining adaptability to varying BS loads.

The throughput of the UE $i$ can then be calculated as:
\begin{equation}
\begin{aligned}
R_i=B_i^w\log_2\left(1+\frac{\sum_{k=1}^{N} u_{i, k} P^{\mathrm{rcv}}_{i,k} }{\sum_{k=1}^{N} s_{i, k} P^{\mathrm{rcv}}_{i,k} +\psi_i}\right)
\end{aligned}~,
\end{equation}
where \( B_i^ {\rm w}=\sum_{k=1}^N u_{i,k}\times U_{\text{PRB},k} \) represents the bandwidth allocated to UE \( i \), and the noise term is given by \( \psi_i=\Gamma\nu B_i^w\xi \). Here, \( \Gamma \) is Boltzmann’s constant, \( \nu \) is the temperature, and \( \xi \) quantifies signal to noise ratio (SNR) degradation. The product \( \Gamma\nu B_i^ {\rm w} \) represents thermal noise, while \( \xi \) captures the noise figure's impact on the \( i \)-th UE throughput.

\subsection{Energy Efficiency}
In a mmWave network, the power consumed by the $j^{\rm th}$ BS, denoted as $ P_{\gNB}^{j}$, is a combination of the power consumed by the Base Band Unit (BBU) and the Active Antenna Unit (AAU). This is described in \cite{gao2021energy} as follows: 
\begin{equation}
       P_{\gNB}^j = \frac{{(P_{{{\text{BBU}}}} + P_{{{\text{AAU}}}} )}}{{(1 - \varrho_{{{\text{cooling}}}} )(1 - \varrho_{{{\text{DC}}}} )}}~,
\end{equation}
where $\varrho_{{{\text{cooling}}}}$ and $\varrho_{{{\text{DC}}}}$ represent the power consumption of the cooling module and DC conversion loss, respectively, while the power consumption of the BBU and AAU are denoted as \( P_{\text{BBU}} \) and \( P_{\text{AAU}} \), respectively. In \cite{gao2021energy}, both BBU and AAU are assumed to consume a fixed amount of power, which is an unrealistic assumption, particularly for \( P_{\text{AAU}} \), as it should be dependent on system load.

To address this limitation, we model the power consumption of the AAU as: 
\begin{equation}
\begin{aligned}
P_{\mathrm{AAU}} =\; & 
\underbrace{N_{\mathrm{RF}}\,N_{\mathrm{CC}}\left(P_{\mathrm{mix}} + P_{\mathrm{ADC}} + P_{\mathrm{DAC}}\right)}_{\text{RF/baseband conversion overhead}}+ \underbrace{N_t\,N_{\mathrm{RF}}\,P_{\mathrm{PS}}}_{\text{beamforming phase‐shifter losses}} 
+ \underbrace{P_{\mathrm{MS}}\,\frac{N_{\mathrm{RF}}}{2}}_{\text{DC power per RF‐chain pair}}\\
&+ \underbrace{N_t\,P_{\mathrm{pa}}}_{\text{static DC bias of }N_t\text{ PAs}}   + \underbrace{\frac{1}{\eta}\,P_{\mathrm{Tx}}^{\mathrm{MAX}}\,\frac{N_{\mathrm{PRB}}^{\mathrm{use}}}{N_{\mathrm{PRB}}}}_{\text{load‐dependent PA power}}\, ,
\end{aligned}
\end{equation}
where \( N_t \) represents the number of antennas, \( P_{\text{pa}} \) is the static power consumption of the power amplifiers (PAs), \( N_{\text{RF}} \) denotes the number of radio frequency (RF) chains, and \( N_{\text{CC}} \) corresponds to the number of carrier components. The power consumption of the mixer, analog-to-digital converter (ADC), and digital-to-analog converter (DAC) are represented by \( P_{\text{mix}} \), \( P_{\text{ADC}} \), and \( P_{\text{DAC}} \), respectively. \( P_{\text{MS}} \) represents the supply power of the radio links, while \( P_{\text{PS}} \) accounts for the power consumption of the phase shifter (PS). The term \( \eta \) denotes the efficiency of the PA, and \( P^{\text{MAX}}_{\text{Tx}} \) refers to the maximum transmission power of the BS. The number of utilized PRBs is denoted by \( N^{\text{use}}_{\text{PRB}} \), while \( N_{\text{PRB}} \) is the total number of available PRBs. The power consumption parameters for different hardware components are taken from \cite{gao2021energy}. This load-dependent power model provides a more accurate representation of energy consumption in mmWave BSs.

EE is a crucial performance metric that quantifies the amount of energy consumed per received information bit. The EE can be mathematically represented as:
\begin{align}\label{EE}
  \EE=\frac{ \sum_{i=1}^{U}R_i}{\sum_{j=1}^{N} P_{\gNB}^{j}}~.
\end{align}
\subsection{QoS}
We propose a QoS-driven SMO framework that dynamically adjusts BS activity while maintaining UE throughput requirements. Unlike latency-based methods, our approach prioritizes throughput-oriented QoS to ensure acceptable data rates under energy-efficient operation. To quantify QoS, we define the UE satisfaction metric as the fraction of UEs that achieve a data rate comparable to an \texttt{All On} baseline:
\begin{equation} U_{\rm QoS}^{\rm satisfied}=\sum_{i=1}^U \mathds{1}_{\hat{R}_i > \alpha_{\text{U}}\bar{R}_i}~, \label{eq:alpha}
\end{equation}
where $\hat{R}_i$ represents the achieved data rate for UE $i$ under the SM strategy, and 
$\bar{R}_i$  denotes the rate under \texttt{All On} strategy, and \( \alpha_{U} \) denotes throughput satisfaction parameter (with \( \alpha_{U} \in (0,1] \)) that serves as a design threshold to determine the minimum acceptable proportion of the full-capacity throughput. Intuitively \eqref{eq:alpha} provides the number of UEs that satisfy the guaranteed bit rate (GBR) requirement. By comparing throughput under SM to the full-capacity state, we establish a fair and adaptive QoS metric that accounts for network dynamics while maintaining acceptable service quality. To ensure acceptable network-wide service quality, we define the QoS satisfaction ratio as the fraction of UEs meeting their guaranteed bit rate (GBR) requirement. This is expressed as:
\begin{equation} \label{QoSB}
\psi_{\text{QoS}}= \frac{U_{\rm QoS}^{\rm satisfied}}{U} > \beta, \quad \beta \in (0,1)
\end{equation}
where $U_{\rm QoS}^{\rm satisfied}$ is the number of UEs satisfying their GBR constraint, and $\beta$ is a service reliability parameter that specifies the minimum acceptable fraction of QoS-compliant UEs. If Eq.~\eqref{QoSB} is satisfied, the QoS constraint is met; otherwise, it is violated. This throughput-based QoS metric provides a practical, application-driven measure of network performance, ensuring BS deactivation maintains acceptable service quality while optimizing energy efficiency.



\section{SMO Framework and Problem Formulation}
This section outlines the overall problem formulation for SMO, reviews state-of-the-art SM strategies, and presents the motivation for MARL to address the limitations of traditional approaches. It also introduces the basic principles of RL to provide context for the proposed MARL framework.

\subsection{Problem Formulation}
The objective of this study is to optimize the transition of BSs into SM to maximize EE while meeting UE throughput requirements as defined by QoS constraints. The operational state of each BS is represented by a binary variable \( m_j \in \{0,1\} \), where $m_j = 1$ indicates that BS $j$ is active, and $m_j = 0$ indicates that it is in SM.
The objective function can than be formulated as:
\vspace{-0.2cm}
\begin{subequations}\label{eq:obj}
\begin{align}\label{eq:obj_a}
\max_{[m_1, \cdots, m_j]}~& \EE \\
\text{subject to}~ 
& \sum_{j=1}^{U} u_{i,j}=1, \quad 
  \psi_{\text{QoS}} > \beta 
\label{eq:obj_b}
\end{align}
\end{subequations}
where \(\EE\) is as in \eqref{EE}, \(u_{i,j}\) as in \eqref{association}, and \(\beta\) as in \eqref{QoSB}. The optimization problem formulated in \eqref{eq:obj} is a complex nonlinear integer optimization problem. Constraint \eqref{eq:obj_b} enforces that each UE is served by a single BS and ensures that the SM strategy satisfies the minimum UE rate requirements. This problem can be categorized as a variant of the multidimensional bin packing problem, which is known to be NP-hard, making it challenging to solve optimally.
\subsection{SOTA SM Strategies}

\subsubsection{Iterative QoS-Aware Load-Based (IT-QoS-LB) SM Strategy}
 
In the load-based approach proposed in \cite{celebi2019load}, each BS load is determined by UE-centric share-based association rather than by the number of connected UEs. Specifically, each UE distributes its load contribution inversely across the number of BSs it can be served by. The load factor of the \(i^{\rm th}\) UE is defined as:
\begin{equation}
L^t_{\text{UE},i} = \left\{\begin{array}{l}
\frac{1}{\sum_{j=1}^N u_{i, j}+s_{i, j}},  \text{ if } u_{i, j} \text{ or } s_{i, j} > 0
 \\
0, \text { otherwise }
\end{array}\right.~,
\end{equation}
where $\frac{1}{\sum_{j=1}^N u_{i, j}+s_{i, j}}$ denotes the number of BSs from which a UE can receive service. This formulation ensures that a UE associated with multiple BSs contributes a smaller share of load to each BS. The total load on the \(j^{\rm th}\) BS is then computed by summing the load contributions from all UEs primarily associated with it:
\begin{equation}\label{load}
L^t_{\text{BS}, j}=\sum_{i=1}^U u_{i,j}L^t_{\text{UE},i}~.
\end{equation}
Based on this load metric, the BS with the lowest load is selected for deactivation to reduce energy consumption. To establish a fair baseline, we adopt the load-based BS selection strategy from \cite{celebi2019load} and extend it into an iterative version, referred to as \texttt{iterative QoS-aware load-based (IT-QoS-LB)} algorithm. In this approach, BSs are sequentially deactivated in ascending order of load. After each deactivation, the QoS impact is evaluated, and if degradation occurs, the algorithm reverts to the previous iteration. This method assumes full knowledge of network load and QoS impact after each deactivation, along with a stationary traffic distribution, implying decisions remain valid over time without adapting to network dynamics. However, in real-world scenarios, such perfect knowledge is unrealistic. The pseudocode for this approach is provided in Algorithm \ref{alg:Load}. Since the load-based approach \cite{celebi2019load} outperforms both the random on/off and wakeup control strategies, we limit our comparison to the load-based scheme.
\begin{algorithm}
\caption{\texttt{IT-QoS-LB} for SMO}
\label{alg:Load}
\begin{algorithmic}[1]
\State \hspace{-0.1cm} Take a realization from the system every \( t_{\text{step}} \) (to be defined in Section.~\ref{proposed}).
\State Measure current load \(\mathbf{L}^t=[L^t_{\text{BS}, 1}, \cdots,L^t_{\text{BS}, N} ]\) using \eqref{load}.
\State Rank BSs load vector \(\mathbf{L}\) in ascending order (lowest load first).
\For{\textbf{each} BS $j$ in ranked order}
        \State \hspace{-0.1cm} Temporarily put BS $j$ into sleep mode.
        \State \hspace{-0.1cm} Compute system-wide QoS metric \( \psi_{\text{QoS}} \).
        \If{$\psi_{\text{QoS}} < \beta$}  \Comment{QoS threshold violated}
            \State \hspace{-0.1cm} Reactivate BS $j$ and revert to previous state.
            \State \hspace{-0.1cm} \textbf{Break} (Stop further BS deactivations).
        \EndIf
\EndFor  
\end{algorithmic}
\end{algorithm}
\subsubsection{All On Strategy}
In the \texttt{All On} strategy, as the name suggests, no BSs are turned into SM, representing a special case where all BSs remain active. This strategy serves as a crucial benchmark for evaluating the effectiveness of other SM strategies. It provides a baseline comparison to assess the impact of SM strategies on network performance.
\vspace{-0.3cm}
\subsection{Motivation and Basics of MARL for SMO}
\subsubsection{A Brief Overview of MARL for SMO}
The optimization problem formulated in \eqref{eq:obj} is inherently non-convex and computationally challenging. Traditional methods often rely on idealized assumptions, leading to suboptimal results in practical deployments. To address this, the EE maximization problem under QoS constraints is formulated as an MDP, defined by the tuple \( \mathcal{M} = (\mtS, \mtA, R, P, \kappa) \), where \( \mtS \) represents the state space, \( \mtA \) denotes the action space, \( R \) is the reward function, \( P \) is the state transition probability function, and \( \kappa \) is the discount factor.  MDP-based problems can be solved using dynamic programming (DP). However, DP methods require well-defined mathematical models, including known state transition probabilities, which are often unavailable in complex and dynamic environments. In real-world wireless networks, factors such as UE distribution and channel conditions exhibit high variability and uncertainty \cite{wu2021deep}, making it impractical to rely on predefined models. To address these limitations, we propose a \texttt{MARL-DDQN} framework that learns optimal SM strategies through model-free, data-driven interaction with the environment. Unlike conventional methods, it requires no prior knowledge of UE distribution or channel conditions, enabling adaptive decision-making under dynamic network conditions.


The fundamentals of single-agent RL are first introduced for clarity, followed by the motivation, implementation, and functionality of the proposed \texttt{MARL-DDQN} approach, detailed in Section~\ref{proposed}.

\subsubsection{Basics of RL}
RL algorithms learn to perform tasks through trial and error. At time \( t \), the agent interacts with the environment by observing the state \( \mathbf{s}^t \), selecting an action \( a^t \), and receiving a reward \( r^t \) based on the action taken. The environment then provides the next state \( \mathbf{s}^{t+1} \). A policy \(\pi\) defines the agent’s strategy for selecting actions based on the current state. The objective is to learn an optimal policy \(\pi^*\) that maximizes the cumulative reward over time \cite{ju2022energy}:
\begin{equation}
    \pi^* = \arg\max_{\pi} \mathbb{E} \left[ \sum_{t=0}^{\infty} \kappa^t r^t \mid \pi \right]~.
\end{equation}

The optimal policy is determined using the Q-value function \( Q^{\pi}(\mathbf{s}^t,a^t) \), which represents the expected cumulative reward for taking action \( a^t \) in state \( \mathbf{s}^t \) while following the policy \( \pi \):
\begin{equation} \label{valuefunc}
    Q^{\pi}(\mathbf{s}^t, a^t) = \mathbb{E} \left[ \sum_{t=0}^{\infty} \kappa^t r^t \mid \mathbf{s}^0 = \mathbf{s}^t, a^0 = a^t \right]~,
\end{equation}
where \(\kappa\) is the discount factor that determines how much future reward contributes to present value. The optimal policy \( \pi^* \) is obtained by selecting the action that maximizes the Q-value function \eqref{valuefunc}. To determine this, the Q-values for all state-action pairs must be available. The optimal Q-function can be derived using the Bellman equation \cite{sutton1998reinforcement}, expressed as:  
\begin{equation}
    Q^*(\mathbf{s}^t, a^t) = r^t + \gamma \sum_{\mathbf{s}' \in S} P_{\mathbf{ss}'}^a \max_{a' \in A} Q^{\pi}(\mathbf{s}', a')~,
\end{equation}
where \( P_{\mathbf{ss}'}^a = P(\mathbf{s}'|\mathbf{s}^t,a^t) \) represents the state transition probability. Determining the optimal policy using the Bellman equation requires knowledge of state transition probabilities, which are challenging to obtain in real-world scenarios. Moreover, the large state space, where UEs can be located anywhere within the SA, and the extensive action space make it computationally infeasible to learn Q-values for all state-action pairs individually. To address this challenge, the Q-learning algorithm, a heuristic approach, was introduced \cite{sutton1998reinforcement}, which  iteratively updates the Q-values \eqref{valuefunc} by adjusting them towards the target value \( Y^t \) to approximate the optimal policy.
\begin{equation}
    Y^t = r^t + \kappa \max_{a'\in \mtA} Q(\mathbf{s}^{t+1}, a')~.
\end{equation} 

This update process resembles stochastic gradient descent, allowing the agent to learn optimal action values through interaction with the environment.

\begin{figure*}[ht!]
	\includegraphics[width=0.9\linewidth]{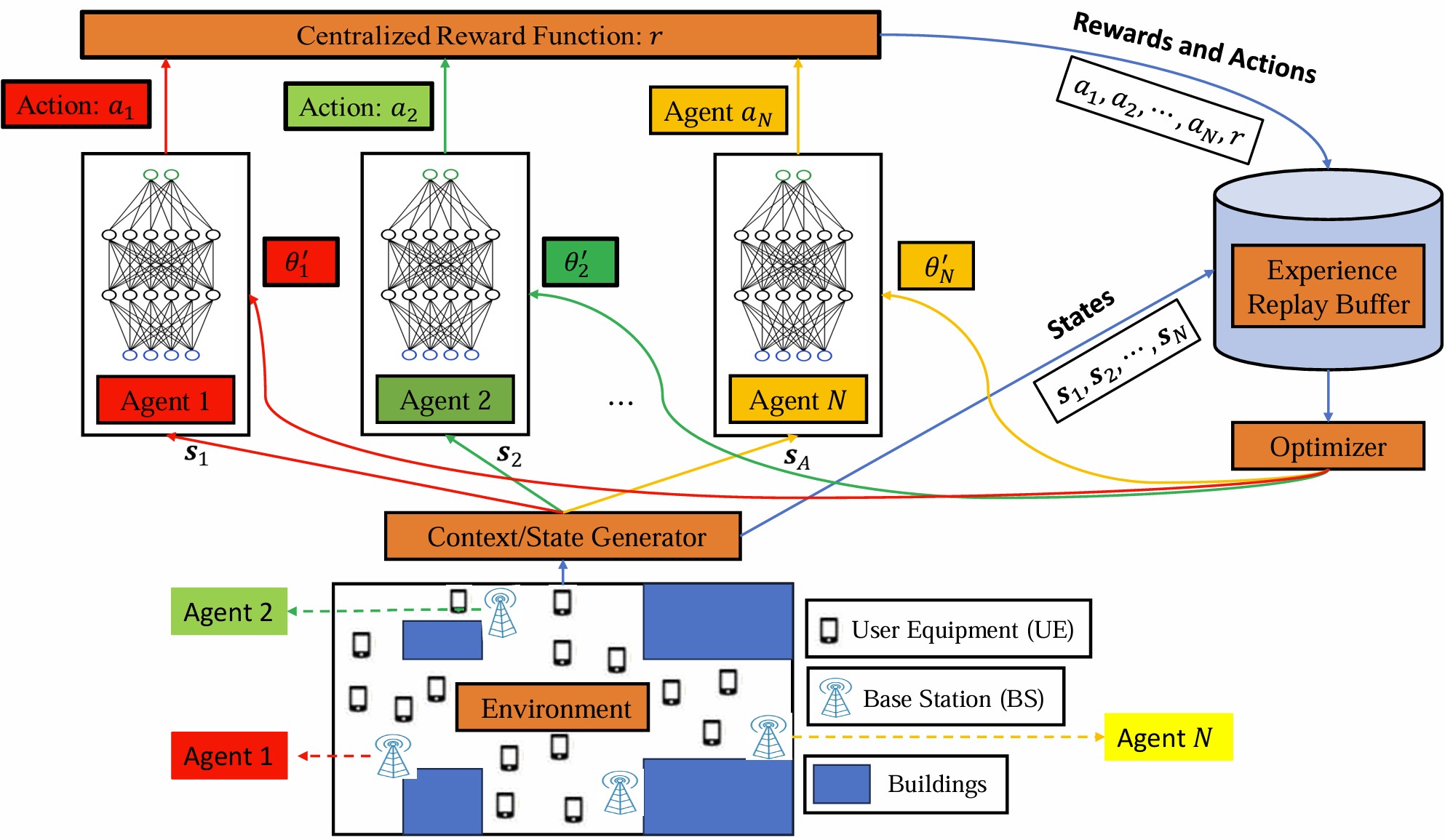}
	\centering
	\caption{ Illustration of the proposed MARL framework for SMO. Each BS is controlled by an independent agent that interacts with the environment and learns a policy for BS SM. }
	\label{MARL}
\vspace{-0.35cm}
\end{figure*}  
\section{Proposed \texttt{MARL-DDQN} Algorithm for SMO} \label{proposed}

To formulate the SMO for mmWave as an MARL task, we represent the system as a combination of five components, as shown in Fig. \ref{MARL}: 1) an environment emulator, which encapsulates the behavior of the BS and UEs; 2) a state generator, which helps the agent in taking action based on the environment; 3) an RL agent, which reacts to the state and takes an action; 4) a reward function, which emits a scalar reward based on the action taken by the agent; and 5) a replay buffer, which serves as the data plane to update the belief of the RL agent.

In the proposed MARL-based SMO framework, each BS is assigned an agent \( b_j \) for \( j \in \{1, \dots, N\} \) responsible for learning its SM policy, as shown in Fig.~\ref{MARL}. At each timestep/iteration \( t \), agent \( b_j \) observes the state \( \mathbf{s}^t_j \), selects an action \( a^t_j \), and receives a reward \( r^t\). These interactions are stored in a replay buffer and used to update the agent’s network parameters \( \mathbf{\theta}^t_j \).


The primary motivation for employing a multi-agent framework is to address the large and complex action space associated with the SMO problem and to reduce the signaling overhead. Additionally, the extensive state space, due to the dynamic and unpredictable distribution of UEs across the SA, further increases the complexity of the problem. A single-agent RL approach would require exhaustive exploration of all possible state-action combinations, making it computationally infeasible. In \cite{masrur2024energy}, SMO was formulated with a fixed number of $N_{\mathrm{off}}$ BSs to deactivate, yielding a combinatorial action space of size $\binom{N}{N_{\rm off}}$.
While this formulation limits the action space to subsets of size \( N_{\rm off} \), it still grows combinatorially with \( N \). However, in this work, the constraint on the number of BSs to be switched off has been relaxed. Instead of predefining \( N_{\rm off} \), the algorithm dynamically determines the optimal number of BSs to transition into SM to maximize EE. This modification expands the action space to:  
\begin{equation}
    \sum_{x=0}^{N} \binom{N}{x} = 2^N~,
\end{equation}  
leading to an exponential increase in the solution space, significantly larger than the fixed-\( N_{\rm off} \) scenario.  

In the proposed distributed MARL framework each agent \(b_j\) operates in a binary action space \( m_j \in \{0,1\} \). While this formulation results in a small action space for individual agents, the overall optimization remains complex. The joint action space across all \( N \) BSs has a cardinality of \( 2^N \), making it computationally intensive to identify the optimal SM configuration.

In a single-agent RL framework, a centralized system must aggregate information on UE distributions, throughput, and QoS requirements across all BSs to determine a SM policy. This necessitates extensive data exchange, resulting in high signaling overhead. In contrast, the proposed MARL framework enables partially decentralized decision-making, where each agent independently selects actions based on local observations and communicates only its decision to the central system for reward computation, thereby significantly reducing signaling overhead while optimizing SMO. Effective coordination among agents is crucial in MARL to avoid suboptimal policies, but it introduces signaling overhead. Furthermore, the environment becomes nonstationary as each BS adapts its policy in response to others, further complicating the learning process. Advanced MARL techniques are required to develop a MARL-based approach that maximizes overall EE without introducing excessive signaling overhead between BSs. This necessitates carefully designing the state representation, action space, and reward function to ensure efficient coordination among agents. The remainder of this section details the proposed algorithm and the formulation of the state-action space and reward function.

\vspace{-0.15cm}
\subsection{DDQN Agent}

When the state and action spaces are large and exhibit complex relationships, tabular Q-learning becomes inefficient and impractical. To address this scalability issue, deep Q-network (DQN) employ neural-network-based function approximation to estimate the Q-values, enabling efficient learning in high-dimensional environments. DQN utilizes a multi-layered NN that maps an \( n \)-dimensional state space to an \( m \)-dimensional action space, with \( \mathbf{\theta} \) representing the network parameters. The primary objective of the DQN is to approximate the optimal Q-function \( Q^*(\mathbf{s}_j^t, a_j^t) \) using a DNN-based function approximator, expressed as \( Q^*(\mathbf{s}_j^t, a_j^t | \mathbf{\theta}_j^t) \).  

However, standard DQN tends to overestimate action values, as it utilizes the same DNN for both action selection and evaluation. To mitigate this issue, we use DDQN, which decouples action selection and evaluation mechanism, maintaining two separate networks: an online network with parameters \( \mathbf{\theta}_j^t \) and a target network with parameters \( \mathbf{\theta}_j^{t-} \). The target network is identical to the online network but is updated every \( \tau \) step to provide more stable target values. The target value for DDQN is defined as  
\begin{equation}
Y^{\mathrm{DDQN},t}_j=r_j^{t}+\kappa\,Q\bigl(\mathbf{s}_j^{t+1},\,\arg\max_{a'\in\mtA}\,Q(\mathbf{s}_j^{t+1},\,a';\,\mathbf{\theta}_j^t);\,\mathbf{\theta}_j^{t-}\bigr)~.
\end{equation}  


The objective is to minimize the loss function \( L(\mathbf{\theta}_j^t) \) to learn the optimal policy \( \pi_j^* \). The loss function for the $j$-th agent is given by,
\begin{equation}
    L(\mathbf{\theta}_j^t) = \bigl(Y_j^{\mathrm{DDQN}, t} - Q(\mathbf{s}_j^t,a_j^t|\mathbf{\theta}_j^t)\bigr)^2~,
\end{equation}  
where the loss formulation represents the mean squared error (MSE) between the target Q-value \( Y^{\mathrm{DDQN},t} \) and the predicted Q-value \( Q(\mathbf{s}_j^t,a_j^t|\mathbf{\theta}_j^t) \). The network parameters \( \mathbf{\theta}^t \) are updated iteratively to minimize this loss, improving the policy estimation over time.

Balancing exploration and exploitation is fundamental in RL. The proposed algorithm employs a time-decaying epsilon-greedy strategy, where each agent selects a random action with probability $\epsilon^{\text{DDQN}}$ and the DDQN-chosen action with probability $1 - \epsilon^{\text{DDQN}}$. The exploration rate decays iteratively as $\epsilon^{\text{DDQN}} \leftarrow \epsilon^{\text{DDQN}} \cdot \epsilon_{\mathrm{th}}^{\text{DDQN}}$, with $0 < \epsilon_{\mathrm{th}}^{\text{DDQN}} < 1$. A higher threshold leads to slower decay and extended exploration, while a lower one accelerates convergence to exploitation. This allows agents to explore more in early stages and gradually exploit learned policies as uncertainty reduces.


A neural network with \( L \) hidden layers is used by each agent to model \( P(r^t_j|a^t_j,\mathbf{s}_j^t, \mathbf{\theta}^t_j) \). Training is performed using the Adam optimizer with L2 regularization (\(\lambda\)). Updating the model after every iteration is computationally expensive and can lead to noisy updates, overfitting, and difficulty in handling concept drift. To mitigate this, the model is updated every \( \tau_{\text{{update}}} \) iterations using a randomly sampled batch of size \( |\textbf{B}| \) from the replay buffer.

 
\subsection{Action Space}  
In the proposed \texttt{MARL-DDQN} framework for SMO, the action space is defined as:
\begin{equation}\label{Action Space}
\mathcal{A} = \left\{ m_j \in \{0,1\} \text{ for all }  j \in \{1,2,\cdots,N \right\}.
\end{equation}
Each agent \(b_j\) makes an independent decision on whether to activate or deactivate its associated BS, aiming to maximize the overall reward.   

\subsection{State Space}  The state is a critical part of the MARL framework, providing essential information for agents to make informed decisions. Each agent receives a distinct state representation, incorporating multiple features. UE distribution is a key feature in state representation, since UEs can be located anywhere within the SA, the state space becomes extremely large, requiring a structured design. To provide relevant UE-specific information while preserving privacy, a clustering-based approach is employed. UEs are grouped into \( K \) clusters at each time step \( t \) using the K-nearest neighbors (KNN) algorithm, similar to Tracking Area (TA)-based grouping, as BSs have access to TA information. The system updates this information whenever a UE transitions to a new TA. Each UE \( i \) is assigned to a single cluster \( k \) (\( k \in \{1, \dots, K\} \)) at each time step, indicated by the binary variable \( \delta^t_{i,k} \), where:  
\begin{equation}
     \delta^t_{i,k} =
   \begin{cases}
     1 & \text{if UE } i \text{ belongs to cluster } k,\\
     0 & \text{otherwise}~.
   \end{cases}
\end{equation}  
This assignment ensures that each UE is exclusively associated with one cluster, satisfying \( \sum_{k=1}^{K} \delta^t_{i,k} = 1 \). The UE distribution across clusters is quantified using the cluster occupation ratio \( \mu^t_k \), defined as:  
\begin{equation}
    \mu^t_k=\frac{1}{U} \sum_{i=1}^U \delta^t_{i,k}~,
\end{equation}  
which indicates the proportion of UEs within each cluster. The overall UE distribution is characterized by the cluster center coordinates \((\tilde{x}^t_k, \tilde{y}^t_k)\) and the corresponding occupation ratio \( \mu^t_k \). The complete clustering information at time \( t \) is represented as:  
\begin{align}
\mathbf{c}^t=[(\tilde{x}^t_1,\tilde{y}^t_1),\cdots, (\tilde{x}^t_K,\tilde{y}^t_K), \mu^t_1,\cdots, \mu^t_K]^T \in \mathbb{R}^{3K,1}~. \label{states}
\end{align}  

The representation in \eqref{states} compactly encodes UE distribution while reducing state space complexity. Next, we incorporate another key factor into the state space: the load factor \(L^t_{\text{BS}, j}\) using \eqref{load}.


In MARL-based SMO, the state representation must capture both spatial and temporal aspects of network dynamics. Accordingly, the state vector for agent $b_j$ at time $t$ is defined as:
\begin{equation}
\begin{aligned} \label{state}
    \mathbf{s}_j^t = \big[ & c^{t - t_{\ell}}, \dots, c^{t-1},
    L^{t-t_{\ell}}_{\text{BS}, j}, \dots, L^{t-1}_{\text{BS}, j},  \psi_{\text{QoS}}^{t-t_{\ell}}, \dots, \psi_{\text{QoS}}^{t-1}, 
    a_j^{t-t_{\ell}}, \dots, a_j^{t-1} \big]^T~,
\end{aligned} 
\end{equation}
where \(t_{\ell}\) denotes the state lookback window. The formulation in \eqref{state} provides a structured representation of historical and real-time network dynamics, enabling informed decision-making. The state \( \mathbf{s}_j^t\) is designed to encapsulate key contextual information, including past UE distribution, load variations, QoS constraints, and agent action history, thereby enhancing the efficiency of the RL policy. The clustering information \( c^{t - t_{\ell}}, \dots, c^{t} \) encodes UE distribution over the past \( t_{\ell} \) time steps. Similarly, the load history at BS \( j \), represented by \( L^{t-t_{\ell}}_{\text{BS}, j}, \dots, L^{t-1}_{\text{BS}, j} \), captures temporal load variations. The QoS performance history \( \psi_{\text{QoS}}^{t-t_{\ell}}, \dots, \psi_{\text{QoS}}^{t-1} \) ensures that the agent prioritizes energy-efficient BS operation while maintaining acceptable QoS levels. Finally, the past actions of agent \( j \), denoted as \( a_j^{t-t_{\ell}}, \dots, a_j^{t-1} \), allow the model to learn correlations between historical decisions and observed network states, facilitating better policy optimization.    

\begin{figure*}[t]
\small
    \centering
\begin{equation}\label{reward}
    r^t =
\begin{cases} 
    \EE, & \text{if } \psi_{\text{QoS}}^{t}> \beta \text{ and } \sum_{j=1}^{N} a_j^t = N  \\  
    \lambda_{\text{QoS}} \cdot \EE \cdot (N-\sum_{j=1}^{N} a_j^t ) - \lambda_{\text{QoS}'}(1 - \psi_{\text{QoS}}^{t}), & \text{if } \psi_{\text{QoS}}^{t}> \beta \text{ and } \sum_{j=1}^{N} a_j^t < N  \\  
    -\lambda_{\text{QoS}'} \left( (1 - \psi_{\text{QoS}}^{t}) + \EE \cdot (N-\sum_{j=1}^{N} a_j^t ) \right), & \text{if } \psi_{\text{QoS}}^{t}< \beta \text{ and } \sum_{j=1}^{N} a_j^t > 0  \\
    -\lambda_{\text{fail}}, & \text{if } \psi_{\text{QoS}}^{t}< \beta \text{ and } \sum_{j=1}^{N} a_j^t = 0  
\end{cases}
\end{equation}
 \noindent\rule{\textwidth}{0.2pt}  
    \vspace{-5mm}  
\end{figure*}

\subsection{Reward Function}
The reward function plays a critical role in MARL, directly guiding policy optimization. In SMO, a well-designed reward is essential to balance energy savings with QoS, as poorly shaped rewards can lead to suboptimal policies. We train the proposed \texttt{MARL-DDQN} algorithm following a Centralized Training and Decentralized Execution (CTDE) framework. Each BS agent independently updates its policy using only its locally observed state–action–reward experiences, without sharing model parameters or replay data with other agents. This ensures that no inter-agent communication or experience exchange occurs during training, thereby minimizing signaling overhead. Although the learning process is decentralized at the agent level, a centralized reward, computed from system-wide energy efficiency and QoS metrics is broadcast uniformly to all agents. This reward formulation aligns individual learning objectives with the global network goal, encouraging cooperative behavior without direct coordination or gradient sharing.

We train the \texttt{MARL-DDQN} algorithm in a decentralized manner to reduce signaling overhead, which would otherwise be required in a centralized approach where agents must share experiences and model parameters $\theta^t$ for all \(j\), leading to massive signaling overhead. However, decentralization leads each BS to optimize independently, often resulting in selfish decisions that harm overall performance. To address this, a centralized reward is uniformly applied across all agents, aligning local actions with global objectives, promoting cooperation without direct coordination.

The reward function \( r^t \)  for each BS is defined in \eqref{reward} as a piecewise function, structured to maximize EE and satisfy QoS constraints. The first two cases in \eqref{reward} apply when QoS requirements are met. If all BSs remain active, the reward is directly proportional to EE, ensuring energy-efficient operation. When some BSs enter SM while maintaining QoS, a higher reward is assigned to encourage energy savings. This is reflected in the term \( (N-\sum_{j=1}^{N} a_j^t) \), which increases as more BSs deactivate, leading to greater rewards. The scaling factor \( \lambda_{\text{QoS}} \) incentivizes BS deactivation, while the penalty term \( \lambda_{\text{QoS}'}(1 - \psi_{\text{QoS}}^{t}) \) prevents excessive energy savings at the cost of degraded QoS. The third case applies when QoS constraints are violated. A negative reward is imposed, increasing in severity as more BSs transition to SM, discouraging excessive deactivation under poor QoS conditions. In the fourth case, if all BSs are in SM and QoS is unmet, a strict penalty of \(\lambda_{\text{fail}}\) is enforced to prevent complete network failure. This reward formulation ensures an adaptive tradeoff between EE and QoS, enabling robust and scalable BS SM control in dynamic wireless networks.

\subsection{Implementation Details of \texttt{MARL-DDQN} Algorithm}

As described in Section \ref{mobility}, each episode is of duration \( T_t\) sec, which consists of two periods: NMP and CMP. The \texttt{MARL-based DDQN} algorithm runs for a total of \( N_{\text{epi}} \) episodes.
At the start of each episode, communities within the SA are randomly redefined to introduce variability and increase the problem's complexity. Instead of keeping communities fixed across all episodes, this approach prevents the model from memorizing traffic patterns and forces it to adapt to dynamically changing UE distributions, enhancing its generalization capability.
During each episode, system realizations are captured at intervals of \( t_{\text{step}} \) and used for training the agents, resulting in a total of \( T_{\text{step}} = T_t / t_{\text{step}} \) steps per episode. The pseudocode for the proposed \texttt{MARL-DDQN} algorithm is detailed in Algorithm.~ \ref{alg:MADDQN}.

\begin{algorithm}
\caption{Proposed \texttt{MARL-DDQN} algorithm for SMO}
\label{alg:MADDQN}
\begin{algorithmic}[1]

\State \textbf{Initialization:}
\State \hspace{-0.1cm} \text{Select \(N\) BS from \(\mathcal{P}_{\rm r}^{\mathrm{BS}}\) (2), and \(U\) UEs.}  
\State \hspace{-0.1cm} Initialize the weight parameters \(\mathbf{\theta}_j\) for all agents \( j \in \{1, \dots, N\} \).
\State \hspace{-0.1cm} Set initial time: \( t = 0 \).  
\State \textbf{Iteration:}
\vspace{0.1cm} 

\For{$e \geq 1$ \textbf{and} $e \leq N_{\text{epi}}$}  
    \State \hspace{-0.1cm} Define the \(C\) communities randomly in SA and their respective areas \(A_n\) and \(A_c\).  
    \For{$t = 1$ \textbf{to} $T_n$} \Comment{Each episode, run $T_n$ steps}  
        \State \hspace{-0.1cm} Run the mobility model as defined in Sec. III-B.  
        \State \hspace{-0.1cm} Take a realization from the system every \( t_{\text{step}} \).  
        \State \hspace{-0.1cm} Construct the state \( \mathbf{s}_j^t \) for all \( j \) using \eqref{state}.  
        \State \hspace{-0.1cm} Obtain action \( a_j^t \) for all \( j \) agents using \eqref{Action Space}.   
        \State \hspace{-0.1cm} Record the reward \( r_t \) as per \eqref{reward}. \hspace{0.1cm} \Comment{Reward is the same for all agents}  
        \State \hspace{-0.1cm} Store all states, actions, and rewards in the replay buffer.  

        \If{$t \bmod \tau_{\text{up}} = 0$} \hspace{0.1cm} \Comment{Update weights every \( \tau_{\text{up}} \) steps}  
            \State \hspace{-0.1cm} Sample a mini-batch of size \textbf{B} from the replay buffer.  
            \State \hspace{-0.1cm} \( \mathbf{\theta}^t_j \gets \mathbf{\theta}^t_j - \lambda_{\text{lr}} \nabla_{\mathbf{\theta}^t_j} L(\mathbf{\theta}^t_j) \). \Comment{Update weights of all agents}  
        \EndIf  

        \State \hspace{-0.1cm} \( t = t + 1 \)  
    \EndFor  
\EndFor  

\end{algorithmic}
\end{algorithm}

\section{Numerical Results and Analysis}\label{sec:Nume}

In this section, we evaluate the efficiency of the proposed SMO approach. The 3D model (Fig. \ref{fig1}), focusing on outdoor UEs, spans an area of $129~{\rm m}\times 206~{\rm m}\times  45~{\rm m}$ ($x, y, z$, respectively), with the UE height set at $1.5$~m above the ground level. The entire area is divided into a grid with a resolution of $1~{\rm m}\times 1~{\rm m}$. The BSs are positioned at the edges of buildings, excluding those near the boundary of the area, resulting in a total of $N_{\rm c}=143$ BS candidate locations, denoted as $\mathcal{P}_{\rm c}^{\mathrm{BS}}$. As explained in Section~\ref{sec:system}, using voxel viewshed algorithm \cite{messerli2015image} we reduce the candidate BS locations to $N_{\rm r}=31$, with locations given by $\mathcal{P}_{\rm r}^{\mathrm{BS}}$ \eqref{RBS}. These BS can provide coverage to the entire SA $\mathcal{P}^{\mathrm{SA}}$ \eqref{PCUE}, which consists of $M=11504$ grid points where a UE could potentially be located. We consider a system operating at $f_{\rm o} = 28$~GHz with transmit power $P_{\mathrm{tx}} = 20$~dBm. Boltzmann’s constant $\Gamma = 1.38 \times 10^{-23}$, temperature $\nu = 298$~K, and noise figure $\xi = 9$~dB. Number of clusters \(C=7\), with \(A_{\rm n}=500\)~m and \(A_{\rm c}=250\)~m. The UE speed varies between \(v_{\rm min}=5\) m/sec and \(v_{\rm max}=5\) m/se

Each episode spanning \(T_{\rm e}=7200\) seconds, comprising \(T_{\rm n}=3600\) seconds and \(T_{\rm c}=1800\) seconds. Each episode consists of \(N_{\rm E}=10\) epochs, with  \(\bar{E}=340\). 
Mobility transition probabilities, such as \(p_{\rm l,c}\), are adopted from \cite{hsu2007modeling}. System realizations occur every $t_{\text{step}} = 360$~s, with the state incorporating $t_{\ell} = 4$ past steps. UEs are clustered into $K = 10$ TAs using K-means. Reward parameters are set as $\lambda_{\text{QoS}'} = \lambda_{\text{QoS}} = 5$ and $\lambda_{\text{fail}} = 20$. Each agent uses a DNN with ($L=4$) of $256$, $196$, $128$, and $32$ neurons, respectively. The Rectified Linear Unit (ReLU) activation function is applied in the hidden layers, while the output layer uses a linear activation function. L2 regularization is incorporated to prevent overfitting, with a parameter of $\lambda=10^{-4}$. For the epsilon decay algorithm, $\epsilon^{\text{DDQN}}=0.7$ is used for the exploration-exploitation trade-off, and the decay rate $\epsilon_{\mathrm{th}}^{\text{DDQN}}$ is set to $0.9$. The model weights $\theta$ are updated after every $\tau_{\text{up}}=4$ iterations, and the batch size is set to $|\mathbf{B}|=256$.

The computational complexity of the proposed \texttt{MARL-DDQN} framework is dominated by neural network inference and Q-value updates.
For a fully connected DNN with (L) layers, (H) hidden neurons per layer, input dimension (D) (state size), and output dimension (A) (number of actions per agent), the per-agent complexity can be expressed as, $\mathcal{O}\big(D \cdot H + (L - 1)H^2 + H \cdot A\big)$,
representing the matrix multiplications across successive layers. For $N = 9$ agents/BS and $K=10$, each DDQN agent receives a state vector of size (D = 132) and outputs Q-values for (A = 2)  resulting in total approximately 1.02 million parameters and 2.04 million floating point operations (FLOPs) per inference step. This linear scalability confirms that the decentralized \texttt{MARL-DDQN} remains computationally lightweight and efficient even for dense network deployments.

\begin{figure*}[t] 
    \centering
    \begin{subfigure}[b]{0.4\textwidth} 
        \centering
        \includegraphics[width=\textwidth]{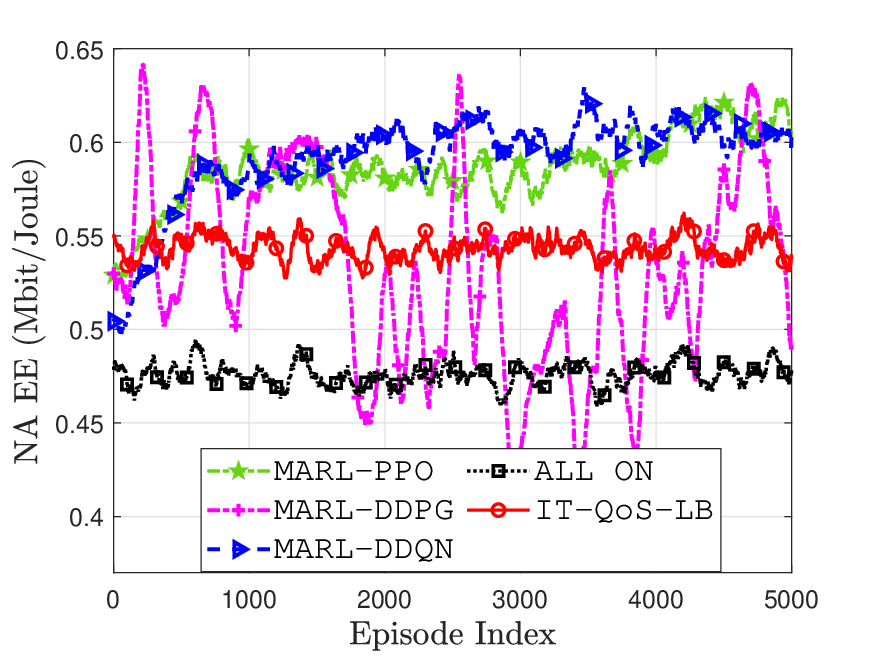}
        \caption{NA EE.}
        \label{EEnUE70}
    \end{subfigure}
    \hspace{1.5mm}
    \begin{subfigure}[b]{0.4\textwidth} 
        \centering
        \includegraphics[width=\textwidth]{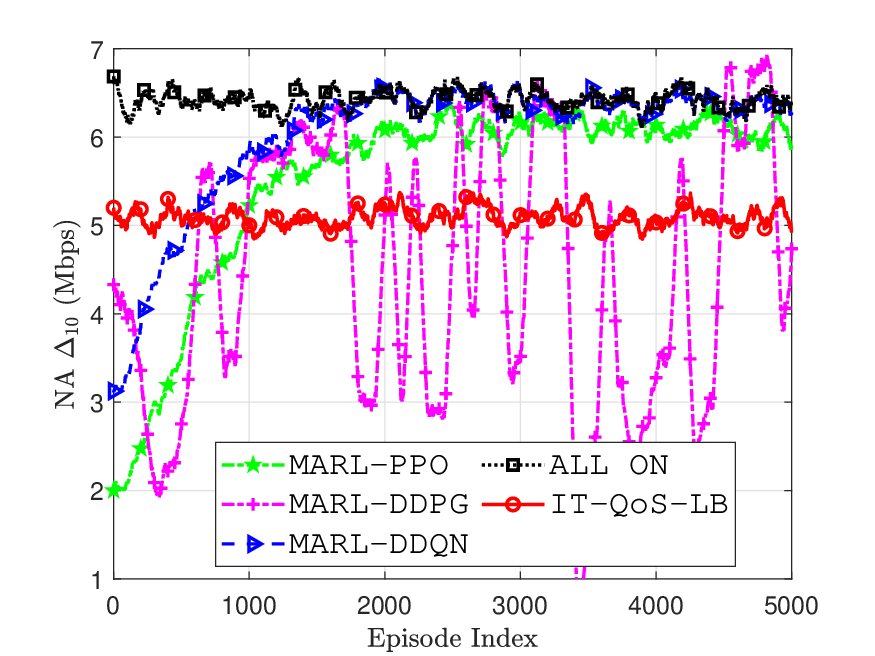}
        \caption{NA \(\Delta_{10}\).}
        \label{10thpnUE70}
    \end{subfigure}
    \vspace{-0.5mm}
    \begin{subfigure}[b]{0.4\textwidth} 
        \centering
        \includegraphics[width=\textwidth]{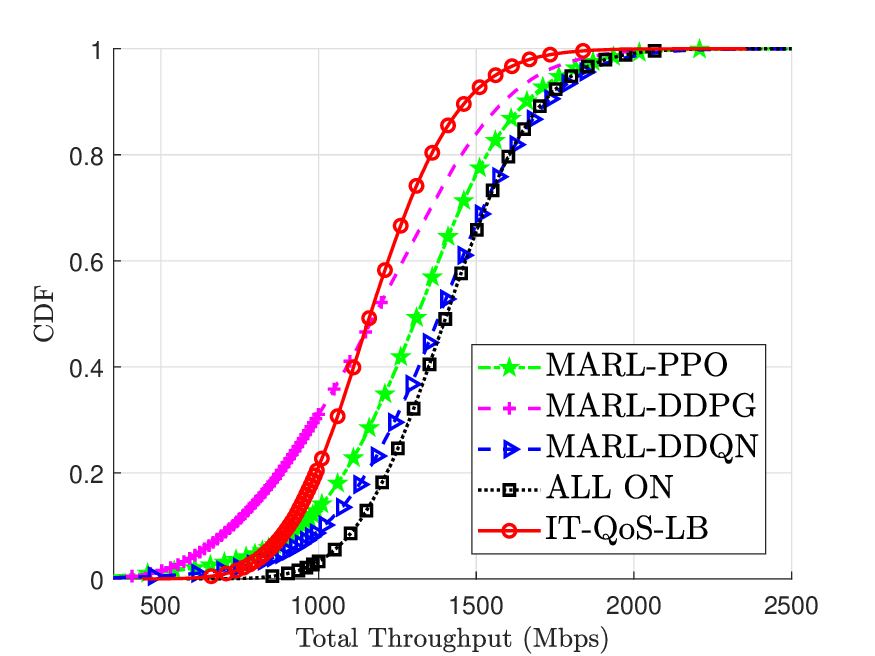}
        \caption{CDF of total system throughput. }
        \label{CDFUE70}
    \end{subfigure}
    \hspace{1.5mm}
    \begin{subfigure}[b]{0.4\textwidth} 
        \centering
        \includegraphics[width=\textwidth]{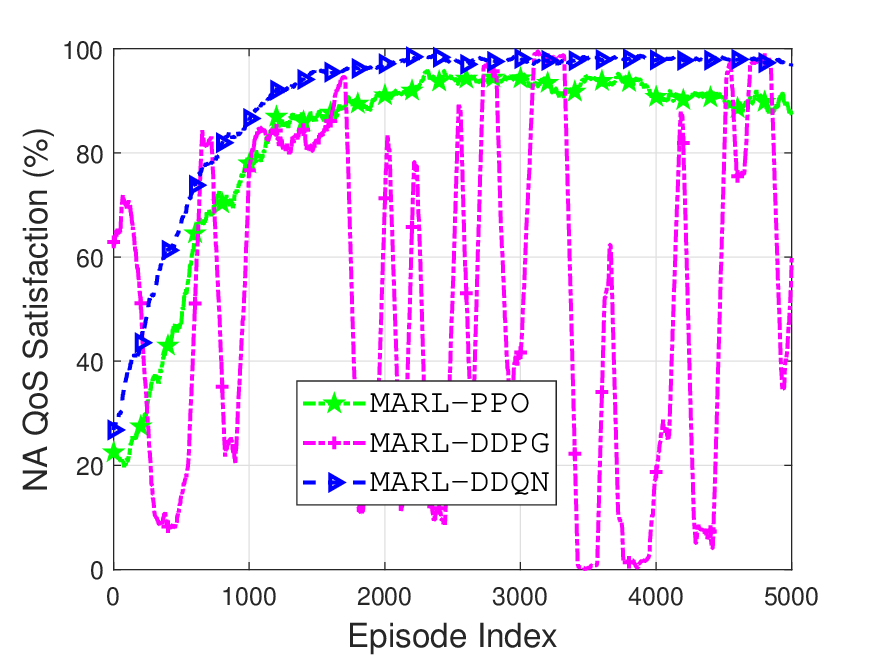}
        \caption{QoS satisfaction percentage. }
        \label{QoSUE70}
    \end{subfigure}

    \caption{Comparison of \texttt{MARL-DDQN}, \texttt{IT-QoS-LB}, and \texttt{All On} strategies over training episodes for \(N=9\) and \(U=70\) in terms of NA EE, \(\Delta_{10}\), total throughput, and the QoS satisfaction percentage. }
    \label{fig:A1}
    \vspace{-5mm} 
\end{figure*}

\subsection{Analysis: Energy Efficiency, Throughput, and QoS}

To evaluate the performance of the proposed \texttt{MARL-DDQN} algorithm, simulations are conducted using \( N = 9 \) randomly selected BSs from a total of $N_{\rm r} = 31$ available BSs, resulting in a large action space of size $32,768$. The number of UEs is set to \( U = 70 \), with parameters \( \alpha = 0.7 \) and \( \beta = 0.7 \), ensuring based on \eqref{eq:alpha}--\eqref{QoSB} that at least 70\% of UEs achieve $70$\% of the throughput achieved by the \texttt{All On} approach is used, presenting a challenging optimization task.

To track the training progress, we compute the Normalized Average (NA) per episode of key metrics (EE, QoS, and 10th percentile (\(\Delta_{10}\))). NA is the average of the metric over all $T_{\text{step}}$, smoothed using a 100-episode moving average to reduce variance and reveal long-term trends. It is defined as:
\begin{equation}\label{NA}
\text{NA}_m^{(e)} = \frac{1}{100 \cdot T_{\text{step}}} \sum_{u=e-99}^{e} \sum_{t=1}^{T_{\text{step}}} m^{(u, t)}~,
\end{equation} 
where \( m^{(u,t)} \) is the metric at episode \( u \), step \( t \). As shown in our prior work~\cite{masrur2024energy}, simpler heuristics (e.g., $\varepsilon$-greedy, UCB, random) show inferior performance. Hence, we retain the strong iterative \texttt{IT-QoS-LB} baseline and focus on advanced learning-based comparisons.

\subsubsection{EE Analysis}
Fig.~\ref{EEnUE70} shows NA~EE. \texttt{All On} yields the lowest EE since all BSs remain active. The iterative baseline \texttt{IT-QoS-LB} achieves competitive EE by sequentially deactivating low-load BSs with QoS rollback checks, benefiting from per-iteration feedback and full observability. Among learning methods, \texttt{MARL-DDQN} converges stably to the highest EE (\(\approx 0.62\)~Mbit/Joule), surpassing \texttt{IT-QoS-LB} and \texttt{All On}. \texttt{MARL-PPO} follows a similar trajectory but requires more episodes to reach comparable EE due to its on-policy nature, which cannot reuse past data and thus needs fresh rollouts per update. In contrast, \texttt{MARL-DDPG} exhibits pronounced oscillations; DDPG is inherently designed for continuous control problems; thus, when it is applied to discrete ON/OFF actions, its Gaussian noise–based exploration becomes ineffective after action thresholding. Furthermore, the tight coupling between the actor and critic amplifies minor estimation errors, leading to unstable learning. This instability is exacerbated by the reward structure of the SMO environment, where abrupt shifts arise from threshold-based QoS penalties and binary base station activation. Since gradient-based methods like DDPG rely on smooth and differentiable reward landscapes to compute stable policy gradients, such discontinuities introduce high-variance gradient estimates, resulting in pronounced training fluctuations. Overall, \texttt{MARL-DDQN} offers the best EE–stability trade-off, with \texttt{PPO} converging more slowly and \texttt{DDPG} remaining unstable.

\subsubsection{10th Percentile (\(\Delta_{10}\)) Analysis}

Fig.~\ref{10thpnUE70} reports NA~\(\Delta_{10}\). \texttt{MARL-DDQN} steadily improves and stabilizes near the \texttt{All On} reference ($6.5$~Mbps), indicating improved worst-user performance while using fewer active BSs (i.e., PRBs). \texttt{MARL-PPO} tracks \texttt{MARL-DDQN} early but lags in convergence due to on-policy sample inefficiency; \texttt{MARL-DDPG} fluctuates substantially, reflecting gradient sensitivity to the discrete, thresholded reward. \texttt{IT-QoS-LB} remains stable yet lower, consistent with its deterministic, load-driven decisions that do not adapt to non-stationary interference as effectively as learned policies. In a dense network, high interference from nearby BSs can degrade signal-to-interference-plus-noise ratio (SINR), particularly for UEs at the cell edge or in high-interference zones, resulting in lower overall throughput. However, turning off a BS alters the interference landscape, leading to reduced interference. 

\subsubsection{Total Throughput Analysis}

The CDF of total system throughput in Fig.~\ref{CDFUE70} shows that the \texttt{All On} strategy yields higher throughput in the lower percentiles (below the 50th), whereas beyond the median, the \texttt{MARL-DDQN} curve closely follows \texttt{All On}. This behavior arises because the lower percentiles correspond to UEs in poor channel conditions or at cell edges. In the \texttt{All On} configuration, every BS remains active, ensuring stronger coverage for these users. Conversely, in \texttt{MARL-DDQN}, selective BS deactivation can temporarily reduce coverage in low-SINR regions, slightly lowering the lower-tail throughput. However, beyond the median, interference reduction and improved load balancing enable \texttt{MARL-DDQN} to match the performance of \texttt{All On} while using fewer active BSs. Quantitatively, the 90th percentile (\(\Delta_{90}\)) total throughput of \texttt{All On} and \texttt{MARL-DDQN} are $1720$ Mbps each, whereas \texttt{MARL-PPO}, \texttt{MARL-DDPG}, and \texttt{IT-QoS-LB} achieve (1672) Mbps, $1604$ Mbps, and $1479$ Mbps, respectively. The comparable high-percentile throughput of \texttt{MARL-DDQN} despite fewer active BSs highlights its ability to balance load and interference through efficient sleep-mode adaptation dynamically. \texttt{MARL-PPO} remains slightly left-shifted due to slower convergence, while \texttt{MARL-DDPG} exhibits the weakest distribution owing to unstable training. The iterative \texttt{IT-QoS-LB} baseline remains competitive yet below the learning methods, highlighting its strength as a heuristic but limited adaptivity under dynamic traffic and interference.

\subsubsection{QoS Analysis}
Fig.~\ref{QoSUE70} presents the NA QoS satisfaction (\%) over training episodes, highlighting how well the proposed \texttt{MARL-DDQN} strategy maintains QoS constraints while optimizing SM decisions. Unlike many existing learning-based SM approaches that primarily focus on EE without evaluating QoS performance, this analysis explicitly evaluates QoS performance, providing insight into policy reliability. QoS satisfaction is calculated as the ratio of time steps within an episode where QoS constraints are met, normalized by the total steps. To mitigate fluctuations, a 100-episode moving average is applied, ensuring a smooth trend. Fig.~\ref{QoSUE70}(d) depicts NA~QoS satisfaction. \texttt{MARL-DDQN} surpasses \(90\%\) by \(\sim\)episode~1200 and stabilizes near \(95\%\text{–}97\%\), demonstrating that the learned policy proactively avoids QoS violations without per-action rollback. \texttt{MARL-PPO} reaches a slightly lower plateau (around \(89\%\)) and requires more iterations, consistent with on-policy updates. \texttt{MARL-DDPG} frequently drops due to unstable actor–critic updates in a discontinuous reward landscape driven by QoS thresholds and binary activation. As expected, \texttt{IT-QoS-LB} enforces QoS by design via rollback, but sacrifices EE and throughput relative to \texttt{MARL-DDQN}.

We also evaluated centralized and decentralized \texttt{MARL-DDQN} configurations. The centralized approach, limited by the exponential joint action space ($2^N$), exhibited unstable convergence and lower EE due to poor exploration and credit assignment. In contrast, the decentralized MARL-DDQN achieved faster convergence and higher EE and QoS.

\begin{figure*}[t] 
    \centering
    \begin{subfigure}[b]{0.32\textwidth} 
        \centering
        \includegraphics[width=\textwidth]{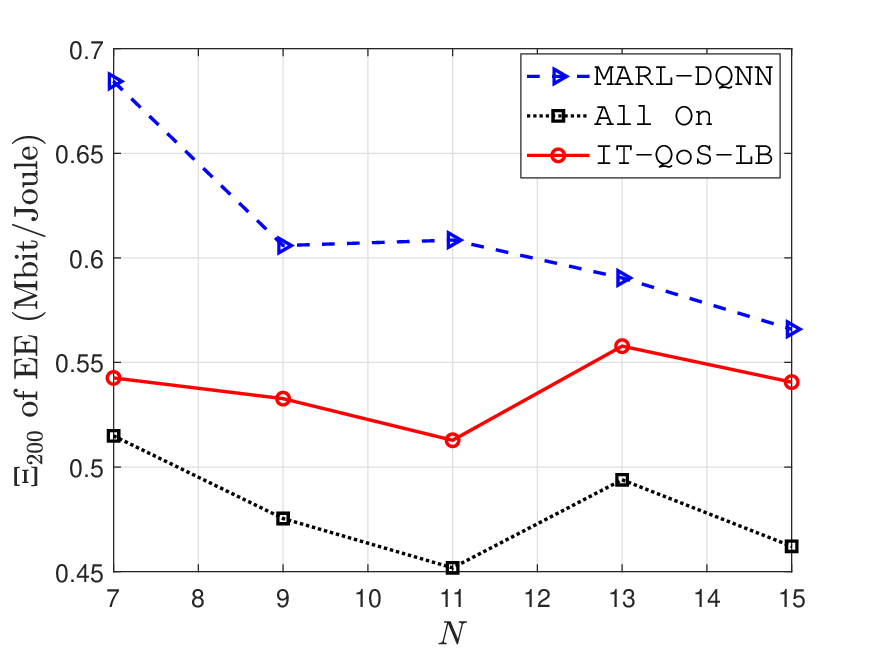}
        \caption{\(\Xi_{200}\) of EE.}
        \label{EEFixUE70}
    \end{subfigure}
    \hfill
    \begin{subfigure}[b]{0.32\textwidth} 
        \centering
        \includegraphics[width=\textwidth]{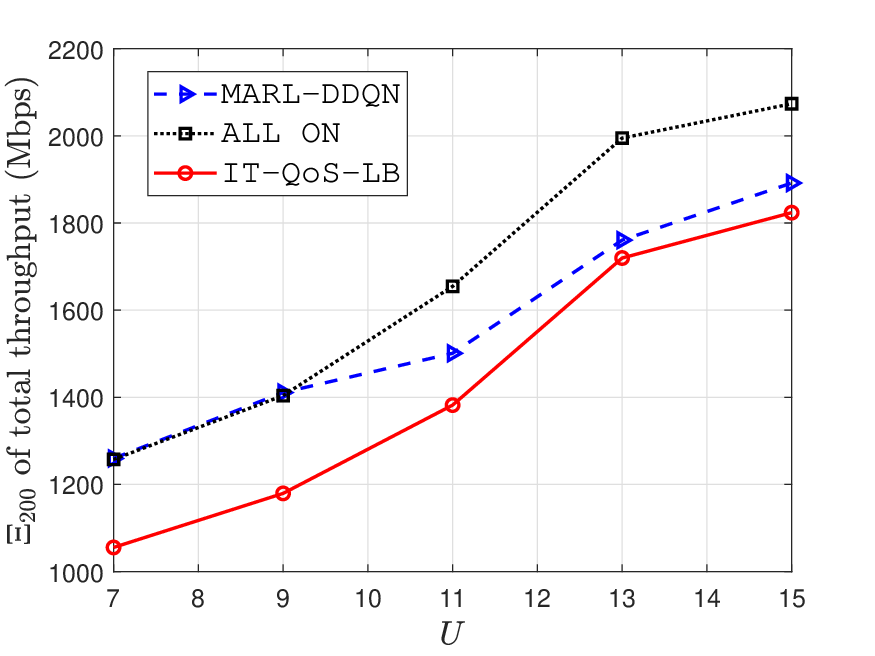}
        \caption{\(\Xi_{200}\) of total throughput. }
        \label{TTFixUE70}
    \end{subfigure}
    \hfill
    \begin{subfigure}[b]{0.32\textwidth} 
        \centering
        \includegraphics[width=\textwidth]{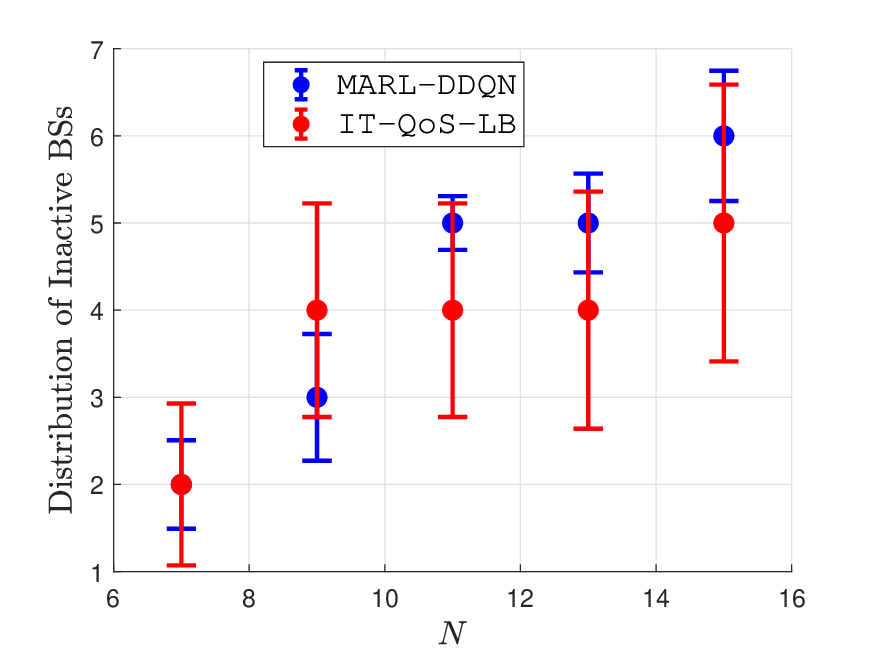}
        \caption{Distribution of BS in SM. }
        \label{BSFixUE70}
    \end{subfigure}

    \caption{Comparison of \(\Xi_{200}\) for EE and total throughput, along with the distribution of BSs in SM over the last 200 episodes, varying the number of BSs (\(N\)) from 7 to 15 while keeping \( U = 15 \) fixed.}
    \label{fig:A}
    \vspace{-5mm} 
\end{figure*}

\subsection{Energy Efficiency and Throughput Versus BS Density}
This section evaluates the scalability of \texttt{MARL-DDQN} for dynamic SMO by varying $N$ from $7$ to $15$ with $U = 70$, using $\alpha = \beta = 0.7$. Performance is assessed in terms of EE and total throughput. The evaluation follows a two-step averaging process: first, the metric is averaged over all time steps within each episode, and then the mean is computed over the last 200 episodes. This metric, denoted as \(\Xi_{200}\) (mean of episode-wise averages over the Last 200 Episodes), provides a stable estimate of system performance after the policy has converged.


\subsubsection{EE Analysis}
Fig.~\ref{EEFixUE70} presents the \(\Xi_{200}\) of EE for the three strategies across different BS configurations (\( N=7\) to \(15 \)). The proposed \texttt{MARL-DDQN} consistently outperforms the baselines, demonstrating its scalability and superior EE. For $N=7$, it achieves an EE of $0.68$~Mbit/Joule, representing a $25.5$\% improvement over \texttt{IT-QoS-LB} ($0.54$ Mbit/Joule) and a $33$\% gain over the \texttt{All On} strategy ($0.52$~ Mbit/Joule). For the \texttt{All On} strategy, a decrease in EE (\(\Xi_{200}\)) is observed as the number of BSs (\(N\)) increases. This trend arises due to two key factors: (1) the total network energy consumption grows proportionally with the number of active BSs, and (2) the increase in BS density leads to heightened inter-cell interference, reducing SE and limiting throughput improvements. Consequently, while the increase in \(N\) marginally enhances total throughput, the disproportionate rise in energy consumption results in a slight but noticeable degradation in EE for the \texttt{All On} approach. 


MARL-DDQN achieves superior EE compared to \texttt{IT-QoS-LB} and \texttt{All On} due to its adaptive interference management. As seen in Fig.~\ref{BSFixUE70}, where the most frequently chosen number of inactive BSs are plotted across different BS densities, \texttt{MARL-DDQN} consistently transitions more BSs into SM as \(N\) increases. The mode and standard deviation of inactive BSs show that \texttt{MARL-DDQN} exhibits lower variability than \texttt{IT-QoS-LB}, indicating stable and consistent BS deactivation decisions. While \texttt{IT-QoS-LB} improves EE through dynamic activation control, its lower deactivation rate and absence of RL-driven adaptability limit its ability to optimize energy savings, resulting in a less efficient SMO strategy compared to \texttt{MARL-DDQN}.


 \subsubsection{Total Throughput Analysis}
Fig.~\ref{TTFixUE70} presents the \(\Xi_{200}\) of total throughput across different BS configurations (\( N=7 \) to \(15 \)). The \texttt{All On} strategy achieves the highest total throughput due to the availability of more PRBs. However, the proposed \texttt{MARL-DDQN} closely follows despite utilizing fewer PRBs, showcasing its efficiency in optimizing resource allocation. As observed in Fig.~\ref{TTFixUE70}, total throughput increases with \(N\) since additional BSs provide more PRBs. The Fig.~\ref{BSFixUE70} shows that the number of inactive BSs increases with \(N\), reducing the available resources. However, the SM strategies, \texttt{MARL-DDQN} and \texttt{IT-QoS-LB}, selectively deactivate BSs, improving SE and helping sustain high overall throughput. Furthermore, \texttt{MARL-DDQN} outperforms \texttt{IT-QoS-LB} in total throughput, demonstrating its ability to adapt to varying BS densities and optimize network resource allocation.



\begin{figure*}[t] 
    \centering
    \begin{subfigure}[b]{0.32\textwidth} 
        \centering
        \includegraphics[width=\textwidth]{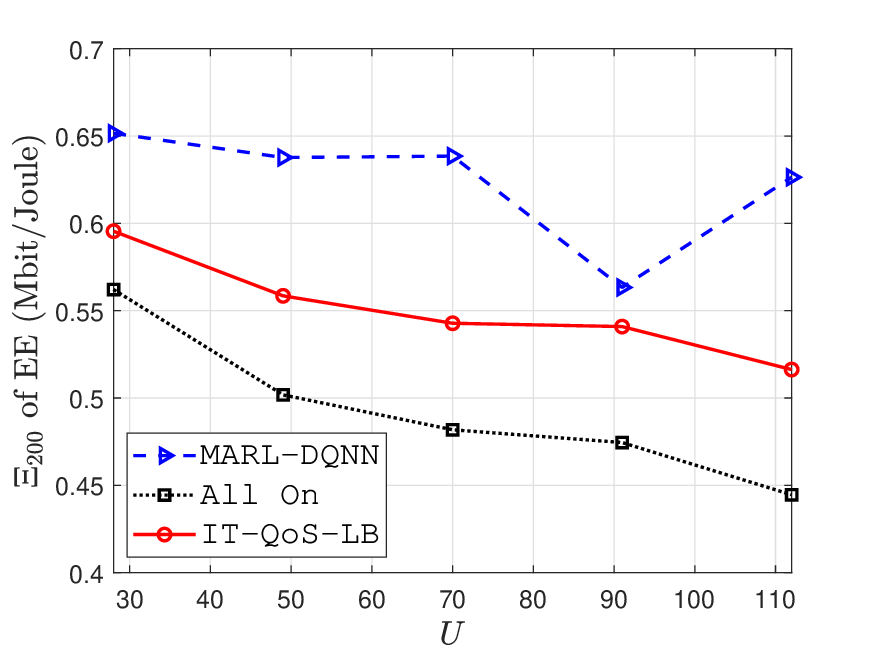}
        \caption{\(\Xi_{200}\) of EE.}
        \label{EEFixBS11}
    \end{subfigure}
    \hfill
    \begin{subfigure}[b]{0.32\textwidth} 
        \centering
        \includegraphics[width=\textwidth]{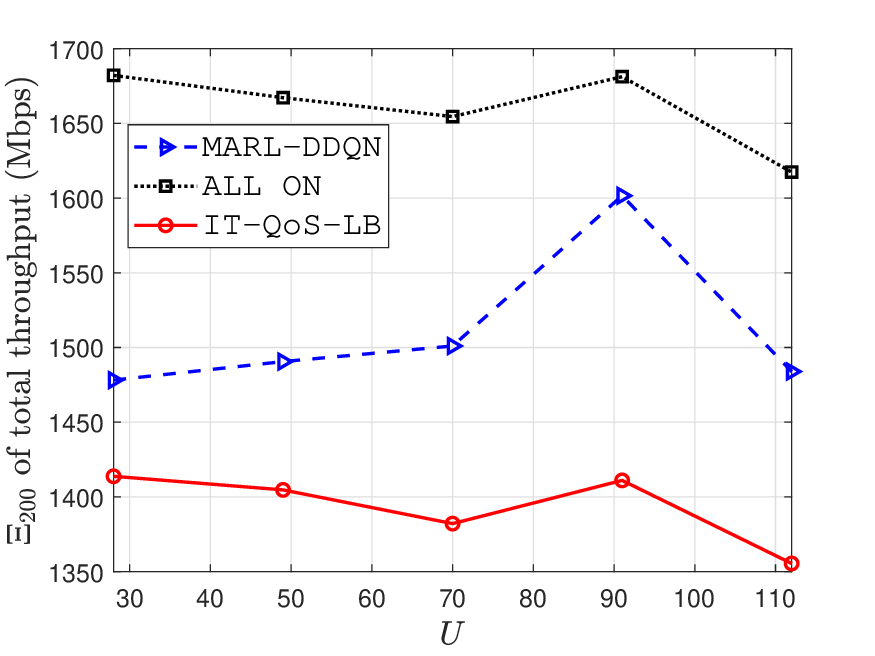}
        \caption{\(\Xi_{200}\) of total throughput. }
        \label{TTFixBS11}
    \end{subfigure}
    \hfill
    \begin{subfigure}[b]{0.32\textwidth} 
        \centering
        \includegraphics[width=\textwidth]{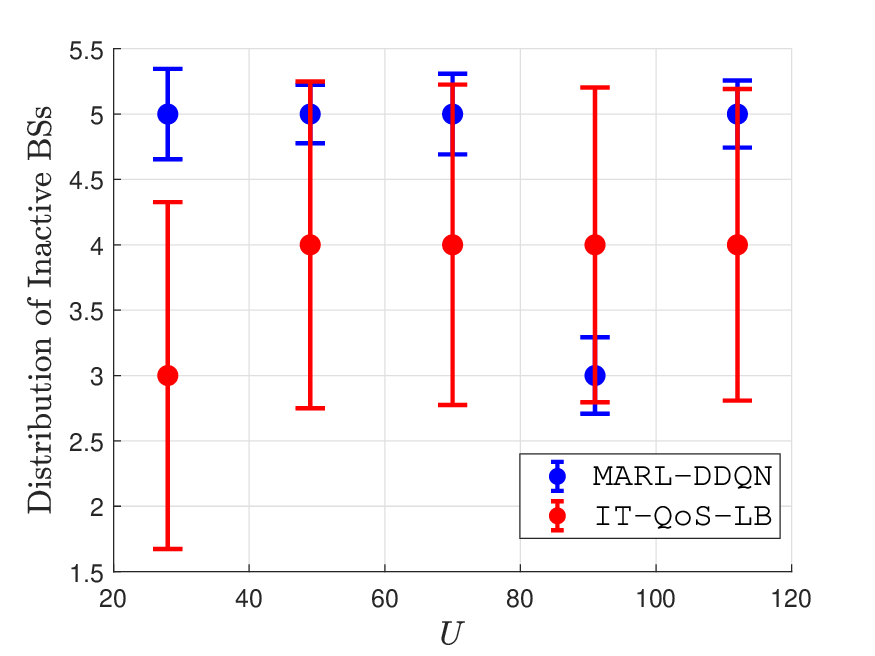}
        \caption{Distribution of BS in SM. }
        \label{UEFixBS11}
    \end{subfigure}

    \caption{Comparison of \(\Xi_{200}\) for EE and total throughput, along with the distribution of BSs in SM over the last 200 episodes, varying the number of UEs (\(U\)) from 28 to 112 with \( N = 11 \) fixed.}
    \label{fig:C}
    \vspace{-5mm} 
\end{figure*}

The proposed MARL-DDQN framework assigns an independent agent to each BS. When a new BS is introduced, a corresponding agent can be initialized and fine-tuned using transfer learning rather than retraining the entire model from scratch. Moderate variations in UE density or mobility can be handled without retraining, though large-scale changes may require additional fine-tuning.

\subsection{Energy Efficiency and Throughput Versus Varying UE Density}

This analysis evaluates the scalability of the \texttt{MARL-DDQN} strategy by fixing \( N = 11 \) while \(U\) from \(28\) to \(112\) with parameters \( \alpha = 0.7 \) and \( \beta = 0.7 \).

\subsubsection{EE Analysis}

Fig.~\ref{EEFixBS11} shows $\Xi_{200}$ of EE for varying UE counts with fixed BSs. As expected, EE decreases with more UEs due to higher BS load, congestion, and increased interference, reducing spectral efficiency. For the \texttt{All On} approach, the decrease in EE is relatively smooth since all BSs remain active, leading to a consistent rise in interference and power consumption.

For \texttt{MARL-DDQN} and \texttt{IT-QoS-LB}, the EE trend is less uniform and follows the variation in inactive BSs, as observed in Fig.~\ref{UEFixBS11}. When $U = 28$, the proposed \texttt{MARL-DDQN} frequently deactivates up to five BSs, resulting in the highest energy efficiency of $0.65$ Mbit/Joule. This represents a $10$\% gain over \texttt{IT-QoS-LB} ($0.59$ Mbit/Joule) and a $16$\% improvement over the \texttt{All On} strategy ($0.56$ Mbit/Joule). As \( U \) increases, more BSs remain active to support QoS, leading to an EE decline. However, for ( U = 91 ), a noticeable drop in EE is observed, as \texttt{MARL-DDQN} activates more BSs and places fewer in sleep mode to meet the increased QoS demand. However, for \( U = 91 \), a noticeable drop in EE is observed, as \texttt{MARL-DDQN} activates more BSs and places fewer in sleep mode to meet the increased QoS demand. Notably, for $U = 28, 49, 70, 112$, \texttt{MARL-DDQN} maintain the same number of inactive BSs. This is attributed to its ability to selectively deactivate different sets of BSs, targeting those that contribute less to overall throughput or cause higher interference. This highlights \texttt{MARL-DDQN}’s capacity to make more strategic SM decisions beyond what is possible with purely load-based heuristics. 

\begin{figure*}[t] 
    \centering
    \begin{subfigure}[b]{0.4\textwidth} 
        \centering
        \includegraphics[width=\textwidth]{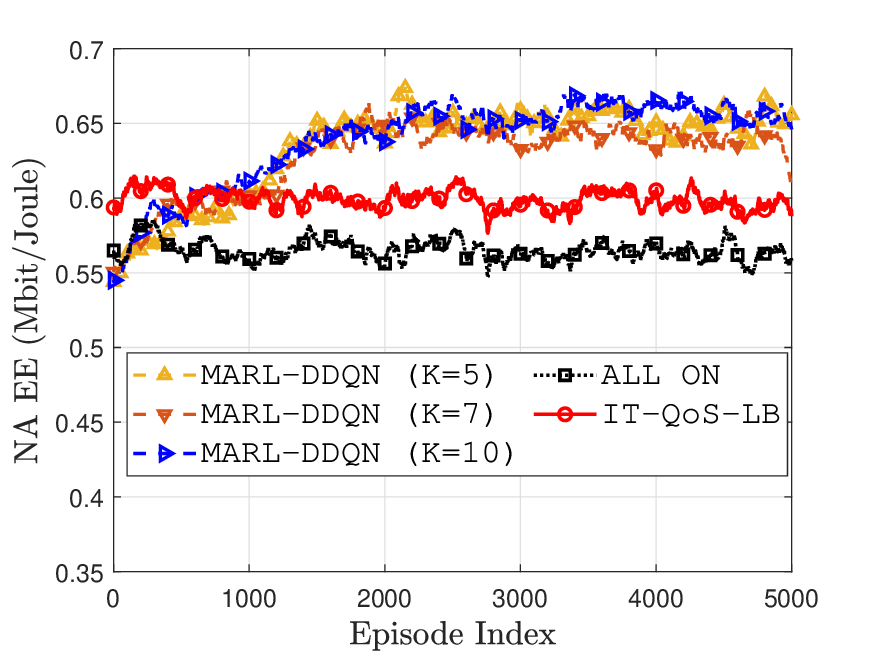}
        \caption{UE density $U = 28$ with varying cluster sizes $K$.}
        \label{VK28}
    \end{subfigure}
    \hspace{1.5mm}
    \begin{subfigure}[b]{0.4\textwidth} 
        \centering
        \includegraphics[width=\textwidth]{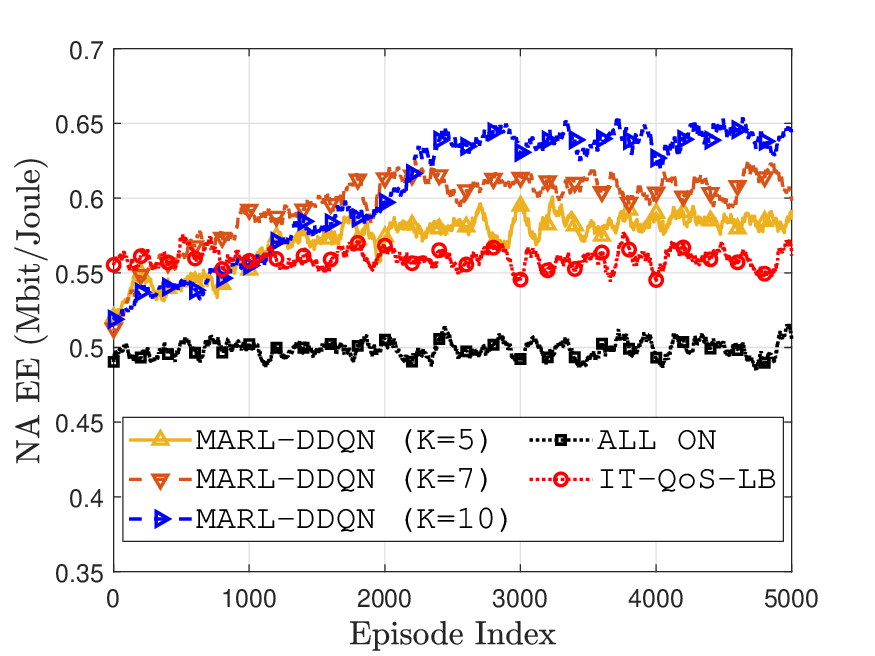}
        \caption{UE density $U = 49$ with varying cluster sizes $K$.}
        \label{VK49}
    \end{subfigure}

    \caption{Impact of cluster size $K$ on normalized average energy efficiency (NA–EE) for varying UE densities.}
    \label{fig:VK}
    \vspace{-5mm} 
\end{figure*}

To study the effect of clustering granularity, we additionally evaluated the impact of varying the number of UE clusters $K$ for two representative UE densities $U=28, 49$ in fig.~\ref{fig:VK}. Results showed that smaller $K$ values perform comparably in low-density scenarios, while higher $K$ (e.g., $K=10$) yields better EEy at higher UE densities by capturing finer spatial variations. For consistency and computational efficiency, $K=10$ is adopted in all main experiments.

\subsubsection{Total Throughput Analysis}
Fig.~\ref{TTFixBS11} presents the \(\Xi_{200}\) of total throughput for varying numbers of UEs. Since the number of BSs is fixed, increasing the number of UEs makes the network more fully loaded, improving overall PRB utilization and leveraging multi-user diversity, thus increasing the total throughput. Although the \texttt{All On} strategy achieves the highest throughput at $U = 91$ with 1681.2 Mbps, it does so at the cost of significantly higher energy consumption. The proposed \texttt{MARL-DDQN} achieves a comparable throughput of $1601.62$ Mbps, while outperforming \texttt{IT-QoS-LB} ($1411.51$ Mbps) by $13.5$\%, demonstrating its ability to sustain high performance through optimized BS activation and effective interference mitigation. The throughput for \( U = 90 \) is higher compared to other cases, as fewer BSs were deactivated, as observed in Fig.~\ref{UEFixBS11}.

To assess generalization, the proposed \texttt{MARL-DDQN} trained with ($N=11$) BSs and ($U=29$) UEs was evaluated on a denser configuration with ($U=70$) UEs without retraining. The model maintained high energy efficiency and ($>95$\%) QoS satisfaction, outperforming \texttt{IT-QoS-LB} and \texttt{All On}. This indicates strong policy generalization across varying UE densities. Since each BS functions as an independent agent, new BSs can be fine-tuned via transfer learning instead of full retraining, which will be explored in future work.

\subsection{Performance Analysis of Varying \(\alpha_{\rm U}\), and \(\beta\)}

This section evaluates the impact of \(\alpha_{\rm U}\) \eqref{eq:alpha} and \(\beta\) \eqref{QoSB} on the performance of \texttt{MARL-DDQN}.

\subsubsection{Impact of \(\alpha_{\rm U}\) }


\begin{figure*}[t] 
    \centering
    \begin{subfigure}[b]{0.4\textwidth} 
        \centering
        \includegraphics[width=\textwidth]{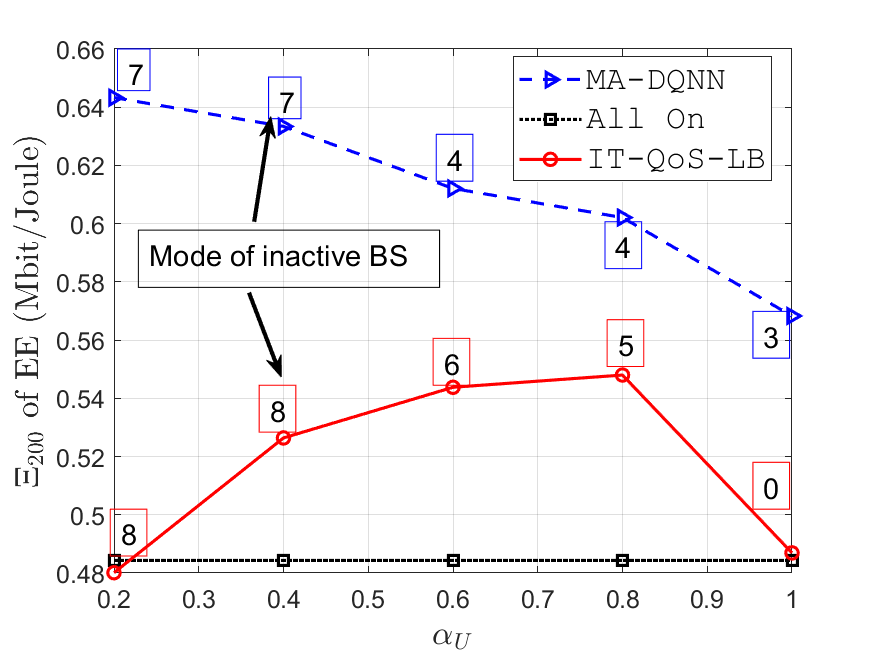}
        \caption{Varying \(\alpha_{\rm U}\).}
        \label{alphavary}
    \end{subfigure}
    \hspace{1.5mm}
    \begin{subfigure}[b]{0.4\textwidth} 
        \centering
        \includegraphics[width=\textwidth]{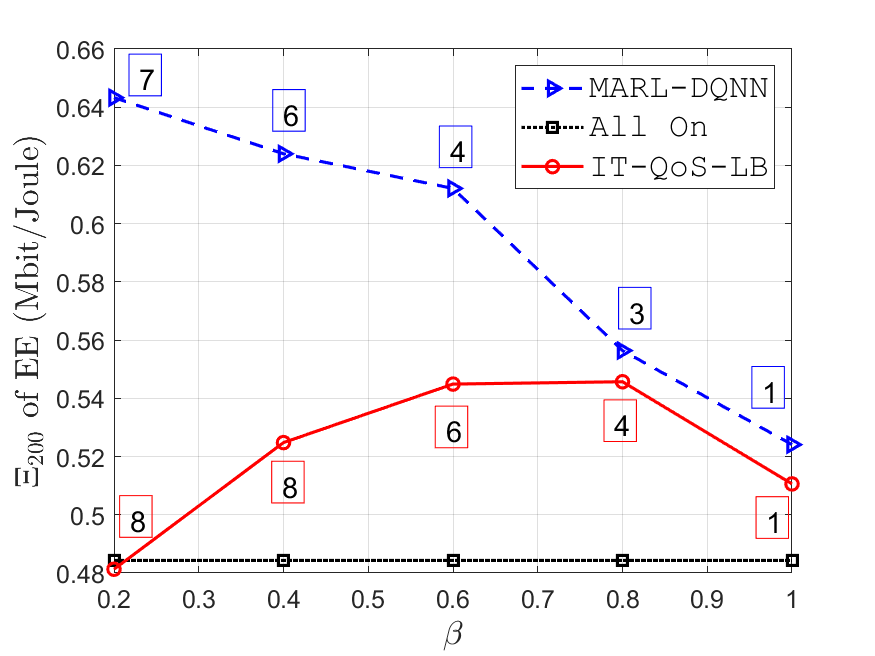}
        \caption{Varying \(\beta\). }
        \label{betavary}
    \end{subfigure}

    \caption{Impact of varying (a) \(\alpha_{\rm U}\) and (b) \(\beta\) on the 200-episode EE metric \(\Xi_{200}\), with fixed \(N=11\).}
    \label{fig:D}
    \vspace{-5mm} 
\end{figure*}

For $\beta = 0.5$, Fig.~\ref{alphavary} shows the impact of varying $\alpha_{\rm U}$ on $\Xi_{200}$ for EE performance. The \texttt{All On} strategy remains unchanged since all BSs remain active regardless of \(\alpha_{\rm U}\). \texttt{MARL-DDQN} tries to maximize EE while meeting QoS. However, as \(\alpha_{\rm U}\) increases, UEs demand higher throughput, prompting \texttt{MARL-DDQN} to keep more BSs active, gradually reducing EE while maintaining a balanced energy–throughput trade-off. The same number of inactive BSs across different \(\alpha_{\rm U}\) values does not imply identical EE, as the policy may deactivate different BSs based on interference conditions, leading to variations in SE. The \texttt{IT-QoS-LB} strategy performs adequately at moderate \(\alpha_{\rm U}\) values but fails to maintain EE at extreme settings. For \(\alpha_{\rm U} = 0.2\), it deactivates an excessive number of BSs (mode of 8), causing network congestion, reduced SE, and lower EE. As \(\alpha_{\rm U}\) increases (0.4–0.8), the BS shutdowns become more balanced, improving power distribution and throughput, leading to increased EE. However, at \(\alpha_{\rm U} = 1\), \texttt{IT-QoS-LB} struggles to determine an effective BS deactivation strategy, leading to all BSs remaining active (mode of inactive BSs = 0). This highlights its lack of adaptive optimization, operating reactively rather than proactively in managing BS activity.

\subsubsection{Impact of \(\beta\)}
In Fig.~\ref{betavary}, we analyze the impact of the \(\beta\) parameter on \(\Xi_{200}\) for EE performance. The \texttt{All On} strategy maintains a constant EE of $0.4843$~Mbit/Joule, as all BSs remain active. \texttt{MARL-DDQN} consistently outperforms \texttt{IT-QoS-LB} across all \(\beta\) values by adaptively optimizing BS sleep decisions while meeting target QoS levels. For instance, at $\beta = 0.2$, it achieves an EE of $0.643$ Mbit/Joule, representing a $34$\% gain over both \texttt{IT-QoS-LB} and the \texttt{All On} strategy. As $\beta$ increases, more UEs must be satisfied, requiring more active BSs. This raises power consumption and reduces EE. For \texttt{MARL-DDQN} the mode of inactive BSs decreases from 7 at \(\beta=0.2\) to 1 at \(\beta=1.0\), confirming that fewer BSs enter SM as QoS requirements tighten.  For \texttt{IT-QoS-LB}, an initial increase in EE is observed from \(\beta\) \(0.2\) to \(0.8\). At low \(\beta\) values (0.2 - 0.4), \texttt{IT-QoS-LB} deactivates too many BSs (mode = 8), overloading the remaining active BSs, reducing SE and thus lowering throughput and EE. As \(\beta\) increases (0.6 - 0.8), \texttt{IT-QoS-LB} retains less BSs in active mode (mode = 6 → 4), improving load distribution. Throughput improves faster than power consumption, resulting in increased EE. However, at \(\beta=1\), \texttt{IT-QoS-LB} excessively activates BSs, causing EE to drop. The non-monotonic EE trend of \texttt{IT-QoS-LB} reflects its reactive and suboptimal BS deactivation strategy.

\section{Conclusion}

This paper presents \texttt{MARL-DDQN}, a MARL framework for energy-efficient BS SMO in mmWave networks under QoS constraints. Unlike traditional SMO approaches based on aggregated traffic or fixed UE distributions, \texttt{MARL-DDQN} dynamically adapts to a realistic, time-varying user mobility model. In contrast to centralized RL methods that rely on full network observability and introduce high signaling overhead, \texttt{MARL-DDQN} enables decentralized decision-making, allowing each agent to learn an optimal policy for its BS. By incorporating a realistic BS power consumption model and accounting for beamforming, the proposed framework enables more accurate EE analysis. Simulation results demonstrate that MARL-DDQN consistently outperforms \texttt{MARL-DDPG}, \texttt{MARL-PPO}, \texttt{IT-QoS-LB} and the \texttt{All On} baselines, achieving higher EE, improved total throughput, and enhanced fairness for worst-case UEs. Additionally, the high QoS satisfaction rate (\(95\)\%) achieved after training confirms the robustness of our approach in dynamically varying environments. The method adapts effectively to varying BS densities, UE distributions, and QoS constraints, demonstrating robust scalability and stability in SM decisions compared to reactive, load-based strategies.
Future work will include the joint optimization of SMO and beamforming to enhance link budgets, along with real-world validation of the proposed EE enhancement approach using the O-RAN XApp framework.

\vspace{-0.2cm}

\bibliographystyle{IEEEtran}
\bibliography{mybib}

\end{document}